\PassOptionsToPackage{pagebackref}{hyperref}
\documentclass[pmlr]{jmlr}
\RequirePackage{graphicx}
\usepackage{booktabs}
\usepackage{mystyle}
\usepackage{longtable}
\usepackage{multirow}
\usepackage{adjustbox}

\makeatletter
\def\set@curr@file#1{\def\@curr@file{#1}} 
\makeatother
\usepackage[load-configurations=version-1]{siunitx} 

\theorembodyfont{\upshape}
\theoremheaderfont{\scshape}
\theorempostheader{:}
\theoremsep{\newline}

\jmlrvolume{252}
\jmlryear{2024}
\jmlrworkshop{Machine Learning for Healthcare}

\title[RDF-MBRL]{Decision-Focused Model-based Reinforcement Learning for Reward Transfer}

\author{\Name{Abhishek Sharma}
       \Email{abhisheksharma@g.harvard.edu}\\ 
       \addr SEAS, Harvard University\\
       Allston, MA, USA
       \AND
       \Name{Sonali Parbhoo}
       \Email{s.parbhoo@imperial.ac.uk}\\ 
       \addr Imperial College London\\
       London, UK
        \AND
       \Name{Omer Gottesman}
       \Email{omergott@gmail.com}\\ 
       \addr Amazon\\
        New York, NY, USA        
        \AND
       \Name{Finale Doshi-Velez}
       \Email{finale@seas.harvard.edu}\\ 
       \addr SEAS, Harvard University\\
       Allston, MA, USA} 

\begin{document}

\maketitle

\begin{abstract}
Model-based reinforcement learning (MBRL) provides a way to learn a transition model of the environment, which can then be used to plan personalized policies for different patient cohorts and to understand the dynamics involved in the decision-making process. However, standard MBRL algorithms are either sensitive to changes in the reward function or achieve suboptimal performance on the task when the transition model is restricted. Motivated by the need to use simple and interpretable models in critical domains such as healthcare, we propose a novel robust decision-focused (RDF) algorithm that learns a transition model that achieves high returns while being robust to changes in the reward function. We demonstrate our RDF algorithm can be used with several model classes and planning algorithms. We also provide theoretical and empirical evidence, on a variety of simulators and real patient data, that RDF can learn simple yet effective models that can be used to plan personalized policies.\footnote{The code for our method is available at \href{https://github.com/dtak/robust_decision_focused_rl_public}{https://github.com/dtak/robust\_decision\_focused\_rl\_public}}
\end{abstract}

\section{Introduction}
Reinforcement learning (RL) is the branch of machine learning where an agent learns to make optimal decisions (with respect to a reward) by interacting with an environment.
One example is an algorithm (agent) that learns to provide treatments (decisions/actions) after observing how the patient's health evolves (environment), with a reward function that penalizes unhealthy patient states.
This evolution of health is determined by the patient's transition dynamics, i.e., how a patient's state will change based on the current state and the chosen treatment. 
For example, it could predict how a patient's blood pressure will change after they are given a treatment like vasopressor.
Model-based RL (MBRL) learns a \emph{model of the transition dynamics} (i.e. a transition model) to plan actions, while model-free RL learns to take actions directly from experience without learning a transition model.
MBRL learns the transition model by predicting the next state given the current state and action. This is also known as maximum likelihood estimation (MLE) because it learns the model that maximizes the likelihood of the observed data.

The reward function is important in RL because it defines the objective of the agent.
However, specifying a reward function can be difficult because there are often multiple objectives to consider, the preferences between objectives can change over time or between patients, and the practitioner may only have a vague idea of what the preferences should be.
Clinicians often have a sense of key objectives they want a reward function to encapsulate, but may want to support different trade-offs when the objectives are competing.
For example, they may want to trade off long- and short-term health-effects or balance the aggressiveness of treatments with other measures of a patient's well-being \citep{lizotte_efficient_2010}
In this paper, we consider a similar setting where the reward function can change---between a \emph{learning phase} and a \emph{deployment phase} (Fig~\ref{fig:fig_1}).
As an example in cancer treatment, during the learning phase, we may learn a transition model based on the data from one hospital that prioritizes maximizing drug efficacy over patient side-effects.
At deployment, however, the learned model may need to be applied in an alternate scenario, e.g., palliative care where the focus may shift to minimizing side effects over maximizing drug efficacy.
Here, the transition model captures the disease dynamics in the body that are expected to be the same across hospitals, although treatment preferences change.
Ideally, we want to learn a transition model that is \emph{robust} to changes in the reward preferences at deployment, where robustness refers to the fact that model performs well across different reward preferences during both the learning and deployment phases.

\begin{figure*}[t]
  \centering
    \includegraphics[width=\textwidth]{figures/fig1_rdf.pdf}
   \caption{Overview of the setting.}\label{fig:fig_1}
\end{figure*}

Learning a transition model using MBRL offers several benefits.
First, MBRL methods are more data-efficient than model-free methods, as they can leverage the model to simulate patient trajectories and plan actions \citep{Deisenroth_2011}.
Second, they allow us to build an understanding of the transition dynamics, which can be useful for understanding how patients respond to treatments.
Finally, they allow reusing the learned model to plan treatments for new patients or when the reward function changes for the same patient.
However, the choice of model class for modeling transition dynamics is important.
While complex model classes (e.g., neural networks) can be expressive, they are difficult to interpret and are prone to overfitting, especially with limited data.
Simple model classes (e.g., linear models) may not model every nuance of the dynamics, but they use less data to train, are more robust to overfitting and can be inspected to understand the learned relationships between variables.

However, using a simple model class introduces errors in the model predictions.
The typical model learning method of maximum likelihood estimation (MLE) minimizes the prediction error but the decision policies learned using such models are no longer guaranteed to be good \citep{joseph_reinforcement_2013}.
Decision-focused (DF) learning is an alternative to MLE which selects the model that maximizes the eventual return of the decisions, rather than just minimizing prediction error \citep{grimm_value_2020}.
This approach has been shown to have better performance than maximum likelihood estimation (MLE) when the transition model is misspecified, as it focuses on the most important aspects of the environment for decision-making \citep{wilder_end_2019,sharma2021learning}.
For example, if the reward function is designed to avoid hypotension, the decision-focused model will focus on predicting the patient's blood pressure accurately if it has to choose between predicting blood pressure and other variables.
However, DF methods are ``overfitting" by learning a model that maximizes the return with respect to the current reward function \citep{wang_learning_2021,nikishin_control-oriented_2022}, which can lead to suboptimal decisions when the reward function changes.
This is a problem in healthcare because the reward preferences of a clinician can change over time.
Prior work \citep{futoma_popcorn_2020} addressed this issue of ``overfitting" by combining the decision-focused learning objective with a maximum likelihood estimation (MLE) objective.
This was done by maximizing the likelihood of the observed data, while also constraining the model to produce high returns under the (fixed) reward function.
Although promising, this approach fails to generalize to new reward functions that are very different from the one used during training.

We address these challenges faced by MBRL for simple model classes by introducing a \emph{robust decision-focused (RDF) learning} objective to learn a simple model that performs well across different reward preferences.
We show that there are several possible models consistent with the decision-focused objective (i.e., they are non-identifiable), and RDF leverages this non-identifiability to learn a model that is robust to changes in the reward function.
RDF trades off the decision-focused objective (optimized to have high returns with the learning-phase reward function) with the averaged decision-focused objective (over a distribution of reward functions expected at deployment-phase).
Our requirements for this reward distribution are minimal: we only require that the practitioner can provide the boundaries of the distribution.
We develop a novel algorithm that allows us to perform such optimization efficiently.
In addition to characterizing the RDF objective, we provide a theoretical analysis that shows RDF can achieve better performance than MLE and DF methods.
We also demonstrate the effectiveness of RDF learning of simple and interpretable transition models for a synthetic simulator, a cancer simulator, and a real-world healthcare dataset for hypotension management.

\subsection*{Generalizable Insights about Machine Learning in the Context of Healthcare}
Model-based RL methods show promise in discovering better treatment policies from historical data, but the sensitivity of these methods to changes in rewards at deployment time poses a key challenge to their adoption in high-stakes clinical domains.
Our main hypothesis is that we achieve robustness across different reward settings by optimizing for average performance while learning a transition model in RL.
Our work provides one of the first solutions to learning transition dynamics models that are robust to these changes and can inform future research on applying MBRL to problems in healthcare.
Our contributions are as follows:
\begin{itemize} 
\item We show that DF learning objective is non-identifiable, i.e, there are multiple models that can achieve the same decision-focused objective.

\item We introduce RDF learning objective that leverages this non-identifiability to learn a model that is robust to changes in the reward preferences, while continuing to have high returns under the learning-phase reward function. We provide a theoretical analysis that shows RDF models can achieve better decision quality than MLE and DF models.

\item On a suite of synthetic and real-world healthcare datasets, we demonstrate that RDF models outperform MLE and DF models in terms of decision-making performance on new reward functions.

\end{itemize}

\section{Related Work}
\label{sec:related_work}
\paragraph{Decision-Focused Model-based Reinforcement Learning.}
Several works have developed algorithms for decision-focused (DF) model-based reinforcement learning \citep{joseph_reinforcement_2013, farahmand_value-aware_2017, wang_learning_2021, nikishin_control-oriented_2022, futoma_popcorn_2020}.
These works focus on building simple models, where the transition model cannot represent the true transition dynamics, that can be used for planning.
In contrast to these, we consider the setting where the learned model must be re-used for different reward functions.

\citet{grimm_value_2020} and \citet{nikishin_control-oriented_2022} note that DF models can be non-identifiable for a given reward function. Both suggest this can be a good thing: if the only evaluation metric is the return of the RL agent (i.e. the DF objective), it is easier to find an optimal solution when multiple equivalent solutions of the DF objective exist. 
However, they do not consider the robustness of different DF solutions to changes in reward function.
In contrast, we show that the DF objective does not optimize for robustness, and may find a solution that is not robust to changes in reward. Furthermore, we exploit this non-identifiability of models with respect to the performance of the agent to find a solution that generalizes better to the changed reward function.

\paragraph{RL with changing objectives.}
Multi-Objective RL (MORL) and reward-robust RL aim to train agents to perform well on multiple reward functions \citep{barrett_learning_2008,lizotte_efficient_2010,mossalam_multi-objective_2016,abels_dynamic_2019,Husain_Robust_2021,derman2021twice}.
The most common setting addressed in the MORL literature involves rewards functions linearly combined using a weight called \textit{preference} \cite{abels_dynamic_2019,mossalam_multi-objective_2016,barrett_learning_2008,yamaguchi_model-based_2019,lizotte_efficient_2010,hayes_practical_2022}.
We make the same assumption.
In both MORL and robust RL, the transition dynamics model (and potentially uncertainty about it) is either explicitly or implicitly (via data) provided as input to the algorithm. The algorithm's goal is to then output a policy that is robust, or can be efficiently recomputed if preferences change.
In contrast, our aim here is to \emph{learn a transition model} which will produce robust policies when optimized for different reward functions.

There are a few exceptions---\citep{wiering_model-based_2014,wan_multi-objective_2021,yamaguchi_model-based_2019} propose MORL algorithms which learn a transition model, but do so (a) by only using $\set{(S_t, A_t, S_{t+1})}$ transitions to learn the model, and (b) by making strong assumptions on the state space (e.g. discrete) or the model dynamics (e.g. SIR model in epidemiology).
Importantly, their model is learned \emph{before} it is used for the MORL step (i.e. model learning does not take into account the policy from the MORL step).
In contrast, our model is learned \emph{along with} the MORL step: we consider which transition model will result in a good policy during the MORL step.

\paragraph{Transfer Learning across Rewards Under Fixed Dynamics.}
The setting of fixed MDP transition dynamics but different reward functions has also been addressed in the transfer learning literature.
The key difference between the transfer learning literature and the multi-objective learning literature in which our work is centered is that our method is not designed to adapt to a new task with new data, but rather find one model which applies well to multiple tasks at once.
\citet{Barreto_Hou_Borsa_Silver_Precup_2020} performs transfer learning in situations where only the reward function differs, by using successor features to decouple a policy's dynamics from expected rewards.
\citet{reinke2021xi} relax some of the assumptions that rewards may be decomposed linearly into successor features for knowledge transfer.
Unlike both of these, our work makes no assumptions about the form of the reward function. The idea of using successor features to express the reward function is complementary and can also be incorporated into our approach.

\section{Preliminaries and Background}\label{sec:background}
\subsection{Notation}
\textbf{Markov Decision Process}
In RL, a Markov Decision Process (MDP) $M$ is defined as $\cM = (\cS,\cA, T_0, R, T_\ast,\gamma)$.
$\cS$ is a state space, $\cA$ is an action space, $T_0$ is the starting-state distribution, $R(s,a)$ is a reward function, $T_\ast(s'|s,a)$ is a transition distribution function, and $\gamma \in [0,1)$ is a discount factor. 
We assume a fixed starting state, although having a distribution over the starting states is straightforward.

The goal of the agent is to learn a policy $A_t = \pi^\ast(S_t)$ that maximises the expected return 
\begin{gather}
\label{eq:expected_return}
J_{T_\ast,R}(\pi) = \E{A_t \sim \pi}{\sum_{t=0}^{\infty} \gamma^t R(S_t, A_t)}
\end{gather}
for transition function $T_\ast$ and reward function $R$.
The policy quality can also be measured by the Bellman optimality error,
\begin{gather}
\label{eq:bellman_error}
L_{T_\ast,R}(\pi) = \sum_{s,a}{\abs{Q^\pi_{T_\ast,R}(s,a) - BQ^\pi_{T_\ast,R}(s,a)}},
\end{gather}
where $Q_{T_\ast,R}^\pi(s,a) \triangleq \E{\pi}{\sum_{t=0}^{\infty} \gamma^t R(S_t, A_t) | s_0 = s, a_0 = a}$ is the action-value function of $\pi$:
and $B$ is the Bellman optimality operator induced by the transition $T_\ast$ and reward $R$ on action-value function $Q$:
\begin{gather}
    BQ(s,a) \triangleq R(s,a) + \gamma \E{T_\ast(s'|s,a)}{\max_{a'} Q(s',a')}
\label{eq:bellman_op}
\end{gather}
The optimal policy $\pi^\ast$ can also be derived from $Q^\ast_{T_\ast,R}$. $Q^\ast_{T_\ast,R}$ is a fixed point of $B$, and can either be obtained by applying the iteration $Q \gets BQ$ in Eqn \ref{eq:bellman_op} until convergence, or by minimizing the Bellman error in Eqn \ref{eq:bellman_error}. Therefore, maximizing $J_{T_\ast,R}(\pi)$ is equivalent to minimizing $L_{T_\ast,R}(\pi)$.

\textbf{Reward Preferences}
We assume a reward function that linearly interpolates between $K$ basis reward functions $R_1$ to $R_k$ according to,
\begin{gather}
    R_w(s,a) = \sum_{k=1}^K w_k R_k (s,a); \quad \sum_{k=1}^K w_k = 1
    \label{eq: rewardfunctiongeneral}
\end{gather}
where $w_k$ denotes a preference for a particular basis, and the $R_k$'s are assumed as given.
In practice, these reward bases are competing reward functions that the practitioner considers important. For example, in cancer treatment, they would promote the treatment's efficacy ($R_1$) and penalize the patient's side-effects ($R_2$).

\textbf{Model-based RL}
Model-based RL (MBRL) algorithms learn a transition model $T_\theta$ of the environment and use this model to plan a policy. Here, $\theta \in \Theta$ are the parameters of the model which need to be estimated.
MBRL methods allow improved sample-efficiency and generalizability \citep{Deisenroth_2011}.
Traditional methods \cite{Sutton_1991,wiering_model-based_2014} use maximum-likelihood estimation to estimate $\theta$ (MLE-MBRL), which is equivalent to minimizing the KL divergence between the transition model and the true dynamics:
\begin{gather}
    \theta_{\MLE} \gets \arg\min_{\theta} \KL{T_\ast}{T_{\theta}}
\end{gather}

Since the MLE objective does not directly optimize for the objective of discounted returns, it can fail to find optimal policies when the model capacity is limited \cite{nikishin_control-oriented_2022,farahmand_iterative_2018,joseph_reinforcement_2013}.
For example, an MLE model may ``waste" its capacity on modeling an action that the optimal policy will never take, at the expense of differentiating between two near-optimal actions because it does not take into account the policies learned with the model.

\textbf{Decision-focused Model-based RL (DF-MBRL).}
DF-MBRL considers the full computational graph of how the transition model $T_\theta$ affects the performance of the policy $\pi^\ast(\theta,R)$:
\begin{gather}
  \theta \rightarrow T_\theta \rightarrow Q^\ast_{T_\theta,R} \rightarrow \pi^\ast(\theta, R) \rightarrow J_{T_\ast,R}(\pi^\ast(\theta, R))
\end{gather}
where 
$J_{T_\ast,R}$ expected return but we can also use the Bellman optimality error $L_{T_\ast,R}$.
The policy $\pi^\ast(\theta, R)$ is the optimal policy for the transition model $T_\theta$ and reward function $R$, and is computed by maximizing $J_{T_\theta,R}(\pi)$. 
$J_{T_\ast,R}(\pi^\ast(\theta, R))$ depends on $\theta$ through its dependence on $\pi^\ast(\theta, R)$. 
We use the notation $J_{T_\ast,R}(\theta)$ or $J_{T_\ast,R}(\pi^\ast)$ to show dependence on $\theta$ or $\pi^\ast$ respectively.
DF-MBRL directly optimizes for the performance of the policy on the true transition $T_\ast$:
\begin{gather}
  \theta_{\DF} \gets \arg\max_{\theta} J_{T_\ast,R}(\theta)
\end{gather}
In settings where the model class of $T_{\theta}$ cannot represent $T_\ast$, the decision-focused model can outperform the maximum-likelihood model \citep{joseph_reinforcement_2013,farahmand_value-aware_2017}.
However, there can be several model parameters $\theta$ whose policy has high performance on the true environment \citep{grimm_value_2020,nikishin_control-oriented_2022}, i.e. the DF model is non-identifiable.

The main limitation of DF-MBRL is that the model is optimized for \emph{one specific reward function}, $R$. While this is not an issue in settings where the reward function is not expected to change, this can be problematic in settings where the reward function is expected to change---or simply not known precisely at training time.

\section{Problem Setting}
\label{sec:problem_setting}
The goal of our proposed approach is to learn a \emph{simple decision-focused model} of the MDP dynamics that produces high-performing policies for rewards encountered during \emph{both} learning and deployment phases. For example, a model for treating patients with a condition in the ICU may need to be deployed in a setting where resources may be more constrained and clinicians must prioritise patient health \emph{as well as minimising costs}. Alternatively, in cancer treatment, different patients during learning and deployment phases may be more susceptible to developing adverse side effects from the treatment and would therefore need lower dosages \citep{lizotte_efficient_2010}.A robust model would perform well across a range of changes in these preferences. 

\paragraph{Learning phase.}
In the \emph{learning phase}, we are given access to the true transition function $T_\ast$ (in the form of a simulator) and the reward function $R_{\bar{w}}$, where $\bar{w}$ is the learning-phase reward preference.
We are also provided a \emph{deployment-phase} reward preference distribution $P(w)$, where $w$ is the deployment-phase reward preference.
During the learning phase, we can simulate trajectories from the simulator without any restrictions.
At the end of this phase, we must build a model of the simulator that is simple enough, but also allows us to plan high-performing policies for different reward preferences.

\paragraph{Deployment phase.}
During the \emph{deployment phase}, we are no longer given access to the true transition function.
Note that while the deployment-phase reward preference $w$ is unknown during the learning phase, we know the distribution $P(w)$ from which $w$ comes from.
In addition to doing well on $R_{\bar{w}}$, we should be able to re-plan high-quality policies on the deployment-phase reward functions.

$\bar{w}$ encodes the preferences for a known patient population, and $P(w)$ would correspond to the set of preferences the doctor wants to support.

\paragraph{Why simple models?}
In healthcare applications, the true environment dynamics are very complex and expensive to access. For example, organs-on-chips models of a lung alveolus can faithfully simulate lung cancer dynamics but are expensive \citep{Ingber_2022, Hassell_lungcancer_organonchip2017}.
For this reason, simple models are learned to understand disease dynamics and to simulate data for reasoning and policy learning.
While complex computational models can be used to simulate highly accurate data, they can be expensive to run and their access can be siloed.
Simple models are desirable for reduced computational complexity, improved interpretability \citep{doshi2017towards}, and to \emph{only} capture the dynamics relevant for the problem.

\paragraph{Choosing $P(w)$.}
Although knowledge of $P(w)$ might seem to be a strong assumption, in real-world settings it is often possible to define a $P(w)$ using reasonable boundary conditions.
For example, in many healthcare applications, patients can express their preferred trade-off between the aggressiveness of the treatment and its side-effects.
A domain expert will both know what kinds of trade-offs are common in their domain, as well as what is the common span of preferences along these trade-offs.
Access to this type of knowledge motivates our use of a uniform distribution over a given range of $w$, encoding the idea that domain experts can often easily provide reasonable ranges of preferences, but not probabilities over the preferences.
Our goal then becomes to be robust over the entire range of reasonable preferences.

\section{Proposed Framework: Robust Decision-Focused Model-based RL}
\label{sec:methods}
While DF-MBRL can learn a model that performs well on the true environment, it is tied to a specific reward function that was used during training.
To alleviate this drawback of DF-MBRL, we leverage the non-identifiability of DF solutions---among the many DF solutions on the learning-phase reward function, we choose the one that would perform well on multiple possible deployment-phase reward functions.

We formalize this notion in the following Robust Decision-Focused (RDF) MBRL objective:
\begin{align}
  \label{eq:RDF}
  \theta_{\RDF} \gets &\arg\min_{\theta} \E{P(w)}{J_{T_\ast,R_w}(\theta)} \notag \\
  & \text{s.t. } J_{T_\ast,R_{\bar{w}}}(\pi^\ast(\theta, R_{\bar{w}})) \geq \delta
\end{align}
which optimizes the model parameters $\theta$ to have high performance on the (yet unknown) deployment reward function $R_w$, while simultaneously achieving high performance on the learning-phase reward function $R_{\bar{w}}$.  Below we first provide a theoretical analysis of this objective and then describe our optimization approach.

\subsection{Theoretical Analysis}
We theoretically characterize the quality of policies achieved by RDF algorithm, and show that our RDF objective can achieve better policies than DF-MBRL in the deployment phase.
Hence, given the same representational capacity, RDF will learn a model that better approximates the optimal Q function across the range of possible deployment rewards.
We provide the proof in the supplement.
\begin{theorem} 
Let $R_{\bar{w}}$ be the learning-phase reward function with preference $\bar{w}$, and $R_w$ be the reward function with an arbitrary preference $w$.
Let $\Qoptw$ be the optimal action-value function for the true MDP for reward function $R_w$.
Let $\Boptw, \Bdfw, \Brdfw$ denote the Bellman optimality operators under the true dynamics, DF model, and RDF model respectively.

Assume $\Qdfw$ and $\Qrdfw$ are fixed points under $\Bdfw$ and $\Brdfw$ respectively.
Further assume that the reward function is bounded, $R_w(s,a) \in [0, \rmax] \forall s, a, w$.

\paragraph{DF Case.}
Consider a DF model trained on $R_{\bar{w}}$. 
If the Bellman operator induced by the DF model achieves the error
$\sup_{s,a} |\Boptwbar \Qdfwbar \sa - \Bdfwbar \Qdfwbar \sa| = \epsdfwbar,$
then
\begin{align}
  \Qoptw \sa - \Qdfw \sa &\leq \frac{\epsdfw}{(1-\gamma)} &\quad \textit{ for $w = \bar{w}$} \\
  \Qoptw \sa - \Qdfw \sa &\leq \gamma\frac{\rmax}{(1-\gamma)^2} &\quad \textit{$\forall w \neq \bar{w}$}
\end{align}

\paragraph{RDF Case.}
Consider the RDF model trained with learning-phase preference $\bar{w}$ and deployment-phase reward preference distribution $P(w)$.
For a $w \in P(w)$, if the Bellman operator $\Brdfw$ induced by the RDF model achieves the error
$\sup_{s,a} |\Boptw \Qrdfw \sa - \Brdfw \Qrdfw \sa| = \epsrdfw,$
then
\begin{align}
\Qoptw \sa - \Qrdfw \sa \leq \frac{\epsrdfw}{(1-\gamma)}
\end{align}
\end{theorem}
For, $w \neq \bar{w}$, the RDF bound is tighter since we explicitly optimize $\epsrdfw$ whereas $\gamma \frac{\rmax}{(1-\gamma)^2}$ is constant.

\begin{figure}
  \centering
  \begin{minipage}{0.63\linewidth}
    \centering
    \includegraphics[width=\linewidth]{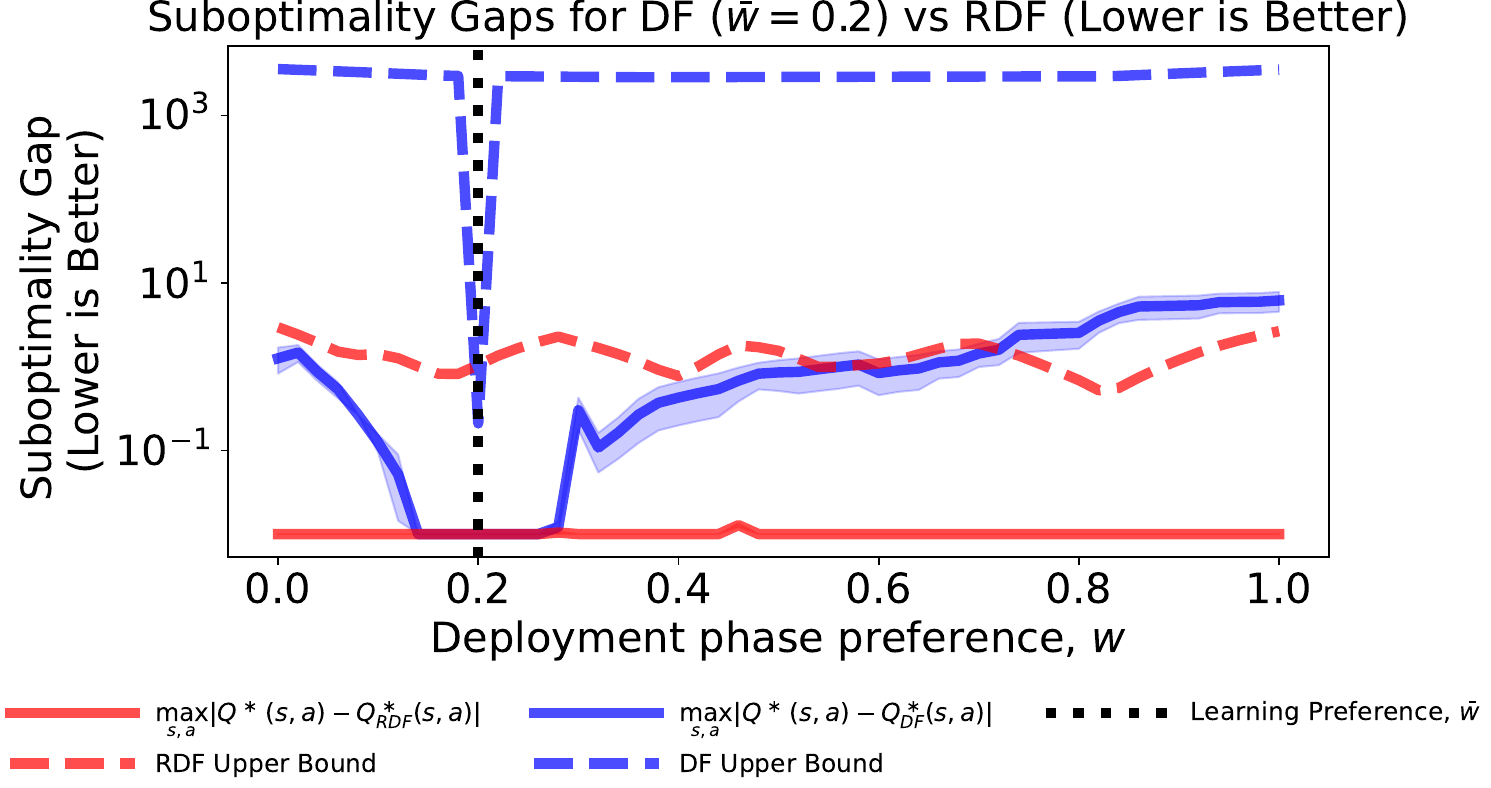}
  \end{minipage}%
  \hfill
  \begin{minipage}{0.35\linewidth}
    \centering
    \includegraphics[width=\linewidth]
    {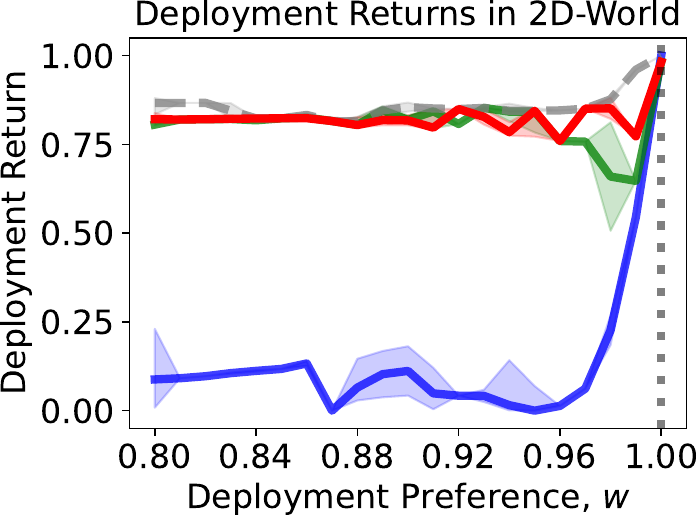}
    \includegraphics[width=0.9\linewidth]{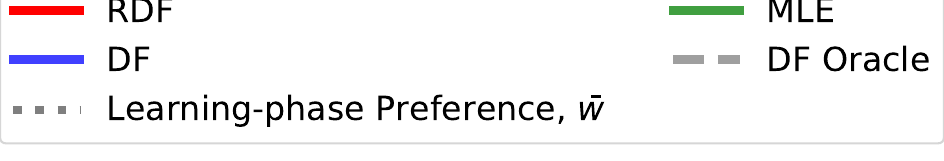}
\end{minipage}
\label{fig:df_vs_rdf_bounds_returns_synthetic}
\caption{\textbf{Left:} Theoretical bounds on the suboptimality gap of RDF models (red dashed line) confirm that RDF models can achieve a better (i.e. lower) suboptimality gap than DF models (blue solid line) away from the learning-phase reward preference $\bar{w}$. Here, the RDF bound guarantees better-than-DF performance for $w > 0.8$. Learning-phase preference $\bar{w}$ is $0.2$ and $P(w)$ is uniform on $[0,1]$. \textbf{Right:} RDF outperforms DF and MLE models by achieving high returns across the deployment preferences in the 2D-World Environment.}
\end{figure}

\paragraph{Empirical validation of Theorem.}
We empirically validate our theorem in the left panel of Figure \ref*{fig:df_vs_rdf_bounds_returns_synthetic} using a simulated MDP with twenty states and two actions (we describe the MDP and provide the simulation code in the supplement).
We plot the suboptimality gaps $\max_\sa |\Qoptw \sa - \Qdfw \sa|$ and $\max_\sa |\Qoptw \sa - \Qrdfw \sa|$ for learning-phase ($\bar{w}$) and deployment-phase ($w$) preferences.
We observe that the RDF bound (red dashed line) is tighter than the DF bound (blue dashed line) for $w \neq \bar{w}$.
\emph{More importantly}, there exist $w$ values for which the RDF bound is tighter than the observed DF suboptimality gap (blue solid line is above the red dashed line in Figure~\ref{fig:df_vs_rdf_bounds_returns_synthetic} (Left)).
For these $w$ values, our bound guarantees that RDF model's suboptimality gap will be lower than the DF model's gap. This is possible because the RDF objective included these $w$ values.


\subsection{Optimizing of RDF objective}
\label{sec:optimization}

Since optimizing the constrained form of RDF objective in Eqn \ref{eq:RDF} can be challenging, we rewrite it using a Lagrange multiplier $\lambda \geq 0$:
\begin{gather}
  \label{eq:lagrangianRDF}
  J^{\RDF}_{T_\ast}(\theta, \lambda) = \E{P(w)}{J_{T_\ast,R_w}(\theta)} + \lambda (J_{T_\ast,R_{\bar{w}}}(\theta) - \delta)
\end{gather}
The limit $\lambda \rightarrow \infty$ recovers the DF objective, while setting $\lambda = 0$ ignores performance on the learning-phase reward and focuses on overall robustness with respect to deployment-phase reward distribution, $P(w)$.

\paragraph{Evaluating the objective.}
As discussed in Sec \ref{sec:problem_setting}, a uniform distribution is a reasonable assumption for $P(w)$ in real-world settings. 
We approximate the expectation in Eqn \ref{eq:lagrangianRDF} using the trapezoidal rule \cite{abramowitz1968handbook}.
Specifically, we construct $\mathcal{W}$, a set of uniformly-spaced $w$ values in the support of $P$ and use it to approximate the expectation:
\begin{gather}
\label{eq:expectation_approx}
    \E{w \sim P(w)} {J_{T_\ast, R_w} (\theta)} \approx \frac{1}{|\mathcal{W}|}\sum_{w \in \mathcal{W}} J_{T_\ast, R_w} (\theta)
\end{gather}
When $w$ is high dimensional, we can use a more appropriate sampling method, and if $P(w)$ is known, we can use importance weights to approximate the expectation more accurately.
The choice of the sampling method is orthogonal to the RDF objective, and we proceed with a uniform grid for simplicity.

\paragraph{Computing the gradient of the RDF objective.}
We employ implicit differentiation to compute gradients through the policy learning step \citep{nikishin_control-oriented_2022}, and use the chain rule to compute the gradient of the RDF objective:
\begin{gather}
  \frac{\partial J_{T_\ast,R}(\theta)}{\partial \theta} = - 
  \underbrace{ \frac{\partial J_{T_\ast,R}(\pi^\ast)}{\partial \pi} }_{\text{Grad Bellman}} \cdot
  \underbrace{ \brackets{ \frac{\partial^2 J_{T_\theta,R}(\pi^\ast) }{\partial \pi^2} }^{-1} \frac{\partial^2 J_{T_\theta,R}(\pi^\ast) }{\partial \pi \partial \theta} }_{\text{Implicit Grad of } \pi^\ast \text{ w.r.t } \theta}
\end{gather}
We can then estimate the gradient of the RDF objective as:
\begin{gather}
\label{eq:gradRDF}
    \frac{\partial J^{\RDF}_{T_\ast}(\theta, \lambda)}{\partial \theta} \approx \frac{1}{|\mathcal{W}|}\sum_{w \in \mathcal{W}} \frac{\partial J_{T_\ast, R_w}(\theta)}{\partial \theta} + \lambda  \frac{\partial J_{T_\ast,R_{\bar{w}}}(\theta)}{\partial \theta}
\end{gather}

\paragraph{Choosing a policy planner.}
Depending on the model parameterization, we can use an appropriate planning algorithm to learn the policy.
The choice of the algorithm is orthogonal to the RDF objective, with the only requirement being that the algorithm can compute the gradient of the policy $\pi^\ast(\theta)$ with respect to the model parameters $\theta$.
However, specific simple model classes support fast planning methods that can be used, e.g. Value Iteration (VI) for tabular transition matrix with discrete states, and Linear Quadratic Regulator (LQR) for linear transition dynamics with continuous states.
For example, in the cancer simulator we describe in Sec \ref{sec:cancer_simulator}, an LQR planner returned an optimal policy in less than a second because we used a linear transition model, while a Deep Q Network (DQN) \citep{Mnih_et_al_2015} takes hours to train.


\begin{algorithm2e}
\caption{General Robust Decision-Focused RL Algorithm}
\label{alg:rdf}
\LinesNumbered
\DontPrintSemicolon
\KwIn{Initial model parameters $\theta$, learning reward preference $\bar{w}$, range for $P(w)$, reward basis functions ${R_1, \dots, R_k}$}
Create uniform grid $\mathcal{W}$ on the range of $P(w)$\;
Initialize Q-function parameters ${\phi_{\bar{w}}}$, $\set{\phi_w: w \in \mathcal{W}}$\;
\Repeat{the Bellman error converges}{
\ForEach{$w \in \set{\bar{w}} \cup \mathcal{W}$}{
Using model $T_\theta$ and reward function $R_{w}$, update action-value function $Q^\ast_{T_\theta,R_{w}} (\phi_w)$\;
Compute policy $\pi^\ast(\theta, R_{w})$ using the action-value function\;
Compute return on true model, $J_{T_\ast,R_{w}}(\pi^\ast(\theta, R_{w}))$\;
}
Update $\theta$ using the gradients computed using Eqn \ref{eq:gradRDF}\;
}
\end{algorithm2e}

\section{Experiments}
We demonstrate the use of the RDF algorithm on a wide variety of simple model classes, such as tabular representation and linear dynamics.
We first evaluate our RDF algorithm on a synthetic environments, where we can control the complexity of the environment and the reward preferences.
Next, we evaluate our algorithm on a cancer simulator, and finally on a real dataset of hypotensive patients in the ICU.
When assuming access to the simulator, we can simulate from the true simulators only during the learning phase and must learn a \emph{simple} model that can be used to plan policies during the deployment phase.

\paragraph{Baselines.}
We compare the performance of our RDF approach to methods that are relevant to our setting. That is, (a) the baseline must output a simple model of transition dynamics, and (b) the model learned using the baseline must transfer to multiple reward preferences.
As identified in the Related Works (Sec \ref*{sec:related_work}), existing methods satisfy one of these requirements, but not both.
We do not compare with model-based MORL methods \citep{wiering_model-based_2014,wan_multi-objective_2021,yamaguchi_model-based_2019} since they do not fit our setting for two reasons: (a) they only perform a single planning step and only \textit{evaluate} the planned policy on different preferences, and (b)
they do not consider the learning of restricted models informed by the MORL step (they just do MLE/Bayesian model learning).
To consider the best one can do with the constraint of a restricted model class, we train a DF-Oracle model that has access to the true transition dynamics during the deployment phase.
Therefore, we compare RDF against DF and MLE (trained once on the training phase setting), as well as DF-Oracle (trained for each deployment phase setting).

\paragraph{Metrics.}
We evaluate the performance of RDF, DF, MLE, and DF-Oracle on the average deployment-phase return, denoted as $J_{avg}$, and the learning-phase return, denoted as $J_{\bar{w}}$.
We scale all returns to $[0, 1]$ by using the maximum and minimum observed returns for a given reward preference in the case of the simulators, and normalize the returns by the behavior policy's return for the hypotension dataset.
This ensures that different weight preferences (with different return ranges) are comparable to each other, and $J_{avg}$ is not dominated by a single $w$ value.
We report the means and standard errors across 10 random seeds.

\subsection{Results on Synthetic Experiment}
\label{sec:synthetic_experiments}

\subsection{2D-World Environment}
We construct a navigational task with two-dimensional continuous states $\in R_{+}^2$ and two-dimensional actions in $\set{ [0,1]^\top, [1,0]^\top }$.
The agent starts at state $[0,0]^\top$ and the episode ends if any of the states crosses value $25$.
The transition model is given by $\textbf{s}' \gets \textbf{s} + \theta \odot \textbf{a}$, where the true model's parameters are $\theta = \theta^\ast = [1.5, 5]^\top$.
For a physical interpretation, the $\theta$ parameters can be thought of as slippage coefficients for the two state dimensions. If $\theta$ is higher, the agent will slip more for an action it takes in that dimension.

\emph{Reward function.}
The first reward basis function $R_1$ only depends on the first state dimension $S_1$. The second reward basis function $R_2$ is only a function of $S_2$.
\begin{align}
  R_1(S) = \begin{cases}
    5, & \text{if } S_1 \leq 2.5\\
    -1, & \text{o/w}.
  \end{cases}
  &;&
  R_2(S) = \begin{cases}
    \frac{20}{17} S_2^2 - 1, & \text{if } S_2 \leq 13\\
    -201 , & \text{o/w}.\nonumber
  \end{cases}
\end{align}
The first reward function incentivizes the agent to stay in the region $S_1 \leq 2.5$, so knowing the correct value of $S_1$ is important to learn a good policy.
The second reward's value increases with $S_2$ until $S_2 = 13$ (a cliff), after which the agent reaches a region of large negative rewards. Without knowing the correct value of $S_2$ it will end up in, the agent would find it hard to learn a good policy.

\emph{Transition model learning.}
We restrict the model class to $\theta =  \theta_c [1, 1]^\top$, where $\theta_c$ is a scalar parameter, which corresponds to a subspace of the full parameter space not containing the true model parameters $\theta^\ast$. 
The environment is challenging because the model forces the agent to either be too aggressive or too conservative in at least one of the state dimensions.
For example, if it believes the slippage in the second dimension is smaller than it actually is (i.e. $\theta_c < \theta^\ast_2$), it will take aggressive steps in the second dimension and will fall off the cliff at $S_2 = 13$.

Since the learning-phase reward only depends on the first state dimension $S_1$, the DF model $\theta_{DF}$ learns a value close to $\theta^\ast_1$ and can ignore the second state dimension $S_2$ without losing performance on the learning-phase reward function.
The RDF model $\theta_{RDF}$ should reasonably estimate slippage in both the first and second state dimensions to achieve high performance.
The MLE model $\theta_{MLE}$ learns the value $(\theta^\ast_1+\theta^\ast_2)/2$ for a uniform random policy.

We set the learning-phase reward preference to $\bar{w} = 1$ (i.e. $R_w = R_1$), and the deployment-phase reward preferences range to $w \in [0.8, 1]$.
We create a grid of $50$ $\theta$ values in the range $[0, 6]$ and compute the optimal policy's return for $\theta$.
We use Fitted Q-Iteration \citep{ernst2005tree} to learn a deterministic policy for any $\theta$. 

\subsection{Conclusions from synthetic domain}

\paragraph{RDF models are more robust to reward function changes.}
Right panel of Fig~\ref{fig:df_vs_rdf_bounds_returns_synthetic} demonstrate the robustness of the RDF models when subjected to reward preferences away from their training settings. Unlike the DF and MLE solutions, RDF achieves near-optimal performance for the model class in both domains, with almost no degradation in learning-phase return.
Performance of the DF model degrades significantly away from the learning-phase reward preference that it was trained on. MLE tends to transfer better than DF in both cases---but not as well as our RDF approach.

\paragraph{RDF offers significant advantages in optimizing trade-offs between objectives.}
The RDF agents consistently reached the goal faster than DF and MLE agents, even as fuel costs increased, all while maintaining reduced acceleration levels.
This suggests that RDF transition models capture the effect of all actions much better than DF and MLE models. 
When transferring to a preference characterized by higher fuel costs, the DF models failed to even reach the goal.

\section{Healthcare simulators}
\label{sec:healthcare_experiments}
Now we apply our RDF approach to two more complex environments relevant to our intended application: healthcare simulators.
\subsection{Healthcare Simulator: Cancer Treatment}
  \label{sec:cancer_simulator}
  The focus of this cancer simulator is on optimizing dosing strategies to reduce mean tumor diameters (MTDs) for patients undergoing chemotherapy for the drug temozolomide (TMZ) \citep{yauney_reinforcement_2018}.
  The domain utilizes a tumor growth inhibition (TGI) model that captures the growth kinetics of diffuse low-grade gliomas (LGG) during and after chemo- and radiotherapy (CRT) of patients \citep{ribba_tumoR_2012}.
  The 5-dimensional states $S_t = (M_t^1, M_t^2, M_t^3, C_t, t)$, consisting of the patient's mean tumor diameters, drug concentration, and current time-step to ensure the Markovian assumption is met.
  The actions are discrete, in $\set{0,1}$, and correspond to whether the drug is administered or not.

  \emph{Reward function}
  The paper introducing this simulator \citep{yauney_reinforcement_2018} included a reward function that can be interpreted as having two components:
  one that promotes a reduction in overall tumor size, and another that penalizes side-effects from using high concentrations of drugs. That is,
  \begin{align}
    R_1(M_t, A_t, M_{t+1}) &= \begin{cases}
      c_1 (M_t - M_{t+1})  + (M_0 - M_T), & \text{if $t = T-1$}.\nonumber\\
    c_1 (M_t - M_{t+1}), & \text{otherwise}.\\
    \end{cases} \\
    R_1(S_t, A_t, S_{t+1}) &= -c_2 C_{t+1}
  \end{align}
  where $M_t = M_t^1 + M_t^2 + M_t^3$ is the total mean tumor diameter at time $t$.
  The parameters $c_1$ and $c_2$ are constants set to $.1$ and $.5$ in the original paper, after observing that these values led to sufficient MTD reduction in the patient population.
  Notably, these parameters do not take into account the range of preferences that a clinician may want to balance in terms of MTD reduction and side-effects.
  This is an important consideration in the real-world settings \citep{lizotte_efficient_2010}.

  \emph{Transition model learning}
  The environment dynamics follow TGI model, which is a system of ordinary differential equations (ODEs) that describe the evolution of the tumor size over time.
  We learn a linear model class to map $(S_t, A_t)$ to $S_{t+1}$ and use the learned model to plan policies for different reward preferences.
  The linear model class and linear reward function allow us to use a Linear Quadratic Regulator (LQR) to plan policies.

We set the learning-phase preference $\bar{w}$ to be $0.75$, and the deployment phase preference range to $w \in [0.5, 1]$.
We employed a Linear Quadratic Regulator (LQR), which leverages the advantages of the linear model class and linear reward function to efficiently plan policies.
We set constraint values $\delta$ in the RDF objective (Eq \ref{alg:rdf}) to $\delta \in \set{0.95, 0.98, 1}$ times the DF return.
We did not try lower $\delta$ values because the unconstrained RDF model gives a high enough learning-phase return of $0.95$ times the DF return.

\begin{figure}[ht]
\includegraphics[width=\linewidth]{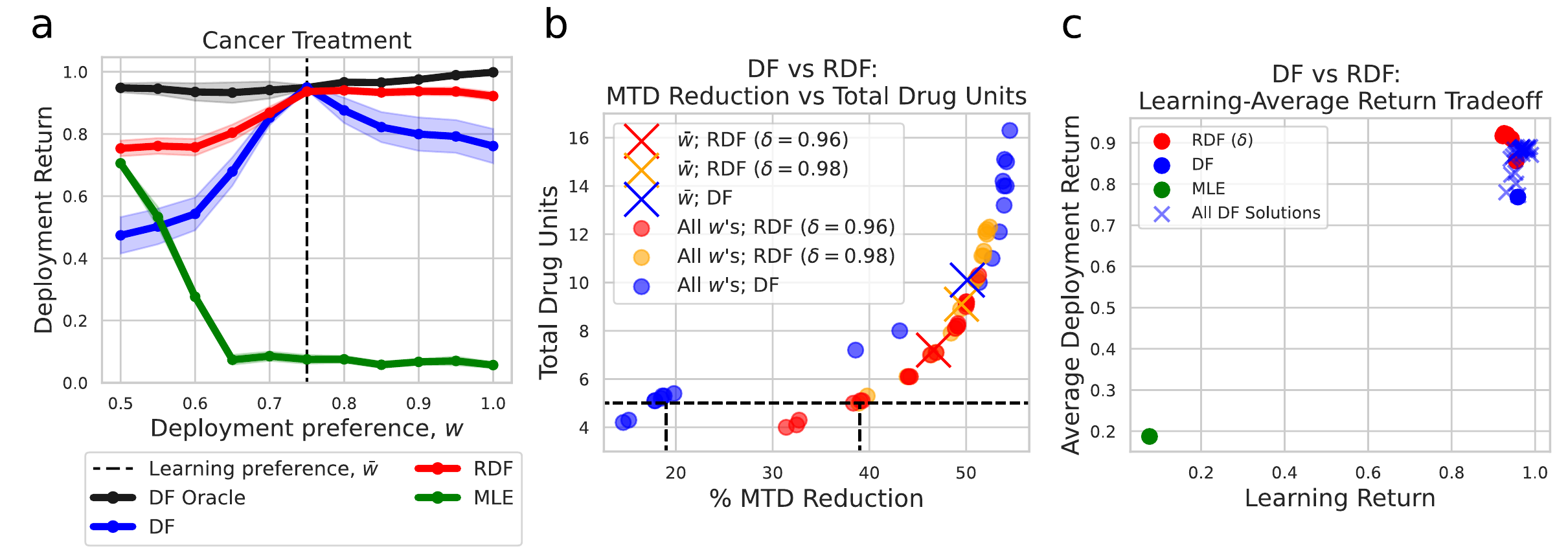}
   \caption{
   Results for Cancer Simulator:
  (a):  
  RDF achieves superior average-case performance and covers DF's performance across preferences for both domains.
   (b): 
   The trade-off between Cancer environment's objectives illustrates how RDF models (red and orange) achieve reductions in tumor diameter, even in low-drug-dose scenarios (they are further right of DF models (blue)). A model achieving a perfect trade-off would be in the bottom-right corner of the plot.
   (c): 
   The learning-vs-average return plot shows how RDF models balance learning and average performance: RDF loses very little in learning return, and gains a lot in average return.
   }\label{fig:fig_cancer_results}
\end{figure}

\subsection{Conclusions from Cancer Simulator}
\paragraph{DF models can be non-identifiable.}
  Fig~\ref{fig:fig_cancer_results}(c) shows that multiple DF solutions (marked by `x' markers) can have very different performance on deployment-time preferences..
  This provides evidence that transfer in DF can be problematic without additional specifications, which RDF provides in form of the average performance objective.

\paragraph{RDF models consistently learn a higher-return policy when evaluated on deployment phase rewards compared to DF and MLE} (Fig~\ref{fig:fig_cancer_results}(a)).
Note that the constraint on the learning return ($\delta$) enables RDF to trade-off learning-phase and average return (Fig~\ref{fig:fig_cancer_results}(c)). Typically, the lower the constraint $\delta$, the higher the average-case performance we can achieve. However, even for the same learning-phase return as DF, RDF model achieves a higher average return.

\paragraph{RDF offers significant advantages in optimizing trade-offs between clinical objectives} (Fig~\ref{fig:fig_cancer_results}(b))
These trade-offs can be particularly relevant when considering individual patient preferences and tolerances.
For patients with low tolerance to side-effects (i.e. treatment constrained to 4-6 doses), RDF demonstrates MTD reduction of approximately 35\%, well above the 20\% reduction by DF.
Even for patients with high tolerance to side-effects, RDF remains competitive, achieving an MTD reduction similar to DF, at 50\%.
These findings underscore the versatility and effectiveness of RDF in optimizing dosing regimens based on patient-specific needs and preferences.
We provide detailed trajectories generated by both RDF and DF models in the supplement. 
\paragraph{RDF models allow learning dynamics using interpretable model classes while still giving good policies}
Choosing a linear model class for transition dynamics allows us to inspect the $(S_t, A_t) \rightarrow S_{t+1} $ relationships that the model has learned.
Table~\ref{tab:cancer_linear_model}(a) presents the linear model's coefficients for predicting the next state's concentration using the current state and action.
We see that the MLE has correctly learned the relationship of concentration ($C_{t+1}$) with the previous concentration ($C_{t}$) and the action of administering a drug ($A_t=1$). It has also learned that the next $C_{t+1}$ is independent of the tumor dimensions ($M_t^1, M_t^2, M_t^2$) and the time-step $t$.
On the other hand, both DF and RDF models have learned non-zero effects of tumor dimensions and time-step on $C_{t+1}$.
Notably, for the RDF model, the effect of $C_{t}$ and $t$ on $C_{t+1}$ is \emph{more} than what the data suggests, meaning that the RDF model is biased to over-estimate the concentration when the existing drug concentration is high or the treatment is in the later stages.
Therefore the optimal policy under the RDF model will hold off giving the drug if there is a penalty for having a high drug concentration, which is a desirable outcome for smaller $w$ values.
The relationship that the RDF model has learned, while not accurate, enables us to find good policy across a range of reward preferences $(w)$ between tumor reduction and drug penalty. 
The DF model learns the opposite effects and thus tends to be more aggressive in its drug dosage.
Note that there is still always a trade-off when the model class is restricted: the MLE model minimized its prediction error at the cost of having a poor-quality policy (Fig~\ref{fig:fig_cancer_results}), and DF/RDF models maximized the quality of their policies at the cost of having higher prediction error (Table~\ref{tab:cancer_linear_model}(b)).

\begin{table}[ht]  
    \resizebox{\columnwidth}{!}{
    \begin{tabular}{lrrr}
    \multicolumn{4}{c}{(a) Linear Model Coefficients for Concentration ($C_{t+1}$) as Outcome}\\
    \toprule
     State/Action & MLE & DF & RDF \\
    \midrule
    Intercept & -0.00 & -0.01 & -0.01 \\
    Concentration $(C_t)$& \textbf{0.76} & \textbf{0.64} & \textbf{0.83} \\
    Proliferative Tissue $(M_t^1)$ & 0.00 & -0.05 & -0.06 \\
    Non-proliferative Quiescent Tissue $(M_t^2)$ & 0.00 & 0.02 & -0.03 \\
    Damaged Quiescent Cells $(M_t^3)$ & 0.00 & -0.10 & -0.01 \\
    Time Step $(t)$ & \textbf{0.00} & \textbf{-0.10} & \textbf{0.13} \\
    Action $(A_t)$& 0.76 & 0.60 & 0.69 \\
    \bottomrule
    \end{tabular}
    \hfill
    \begin{tabular}{llll}
    \multicolumn{4}{c}{(b) RMSE for all Outcomes}\\
    \hline
    Outcome    & MLE   & DF    & RDF \\ \hline
    $C_{t}$ & 0.142 & 3.936 & 1.699 \\
    $M_t^1$ & 0.873 & 3.476 & 2.358 \\
    $M_t^2$ & 6.804 & 7.527 & 7.412 \\
    $M_t^3$ & 5.193 & 5.917 & 5.298 \\ \hline
    \end{tabular}
    }
    \caption{Cancer Simulator: (a) Coefficients for current state and actions for the linear model predicting the Concentration of the next state. MLE learns a zero effect of time step on concentration, whereas DF and RDF models learn nonzero and \emph{opposite} effects. (b) RMSE of predicting the next state given the current state and action.}
    \label{tab:cancer_linear_model}
\end{table}

\section{Real ICU Data - MIMIC IV}

\subsection{Data}
\paragraph{Cohort Selection}
We use EHR data from the MIMIC-IV care database, which contains deidentified clinical data of patients admitted to the Beth Israel Deaconess Medical Center ICU unit \citep{johnson2023mimic}
We filter the data to include only patients who were admitted to the ICU for a total of at least 24 hours, were aged 18-80 years, and had at least 7 mean arterial pressure (MAP) measurements of less than 65 mmHg within the first 72 hours of their ICU stay. We set the 72-hour limit because the care for hypotension during later periods of ICU stay may be quite different.
After filtering, we were left with 5939 stay IDs. We randomly split this dataset $\mathcal{D}$ into a training set $\mathcal{D}_{\text{train}}$ of 2000 stay IDs, a validation set $\mathcal{D}_{\text{val}}$ of 1500 stay IDs, and a test set $\mathcal{D}_{\text{test}}$ of 2439 stay IDs.
\paragraph{Clinical Variables}
Given the hypotension-related cohort selection, we select the following variables for our model: mean arterial pressure (MAP), Glasgow Coma Scale (GCS), PaO2/FiO2 ratio, creatinine, vasopressor use, and fluid bolus use. Among these, vasopressor use, and fluid bolus use are our action variables, and the rest inform the patient state. We impute missing values by first forward filling, then backward filling, and finally filling with the median value of the variable.

\begin{figure}[ht!]
    \centering
    \includegraphics[width=.9\textwidth]{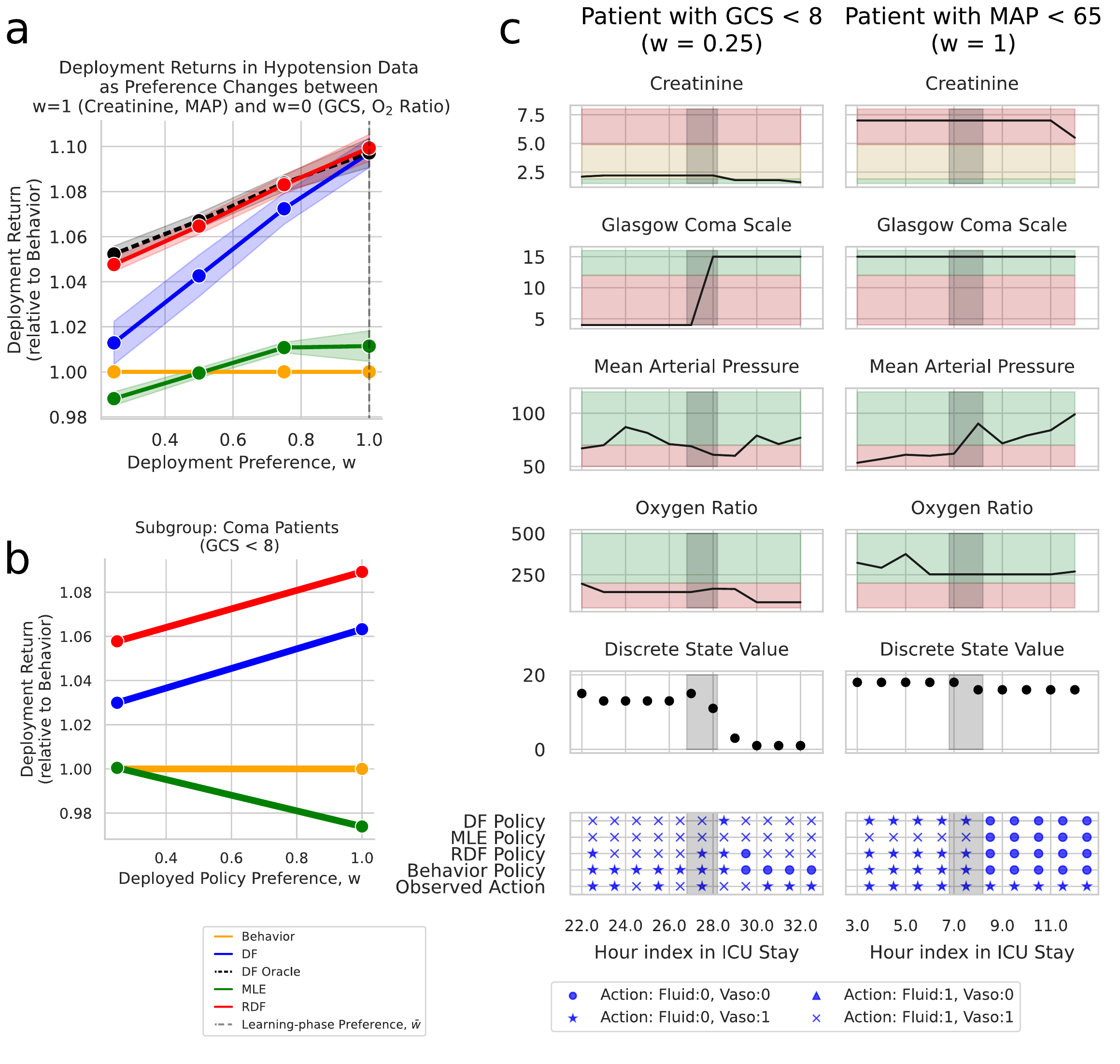}
    \caption{Performance comparison of RDF vs baselines on MIMIC Hypotension data. (a) RDF is better able to achieve higher deployment returns across a range of different deployment preferences across hypotensive patients. (b) RDF consistently outperforms baselines in terms of deployment returns for both acutely hypotensive and comatose patients. For acutely hypotensive patients, DF and RDF baselines prioritise MAP to achieve higher returns than MLE. For comatose patients, both these solutions prioritise improving GCS to achieve higher returns than MLE. RDF manages performance over varying preferences more efficiently than DF. (c) For comatose patients, the RDF policy matches the behaviour policy of the clinician that suggests using only vasopressors which increases the GCS score in the window of interest. For acutely hypotensive patients, DF and RDF policies both suggest using only vasopressors which improves the mean arterial pressure in the window of interest.}
    \label{fig:mimic_results}
\end{figure}

\subsection{MDP Details}

\paragraph{State- and action-space construction}
We discretize the continuous variables into bins that correspond to severity levels. The state space corresponds to all possible combinations of these severity levels. The discretization is done as follows:
\begin{align*}
    \text{O2} = \begin{cases}
        0 & \text{if } \frac{\text{PaO2}}{\text{FiO2}} \geq 200 \\
        1 & \text{if } \frac{\text{PaO2}}{\text{FiO2}} < 200
    \end{cases}&\quad
    \text{MAP} = \begin{cases}
        0 & \text{if } \text{MAP} \geq 70 \\
        1 & \text{if } \text{MAP} < 70
    \end{cases}\\
    \text{GCS} = \begin{cases}
        0 & \text{if } \text{GCS} \geq 12 \\
        1 & \text{if } \text{GCS} < 12
    \end{cases}&\quad
    \text{Creat} = \begin{cases}
        0 & \text{if } \text{Creatinine} \leq 1.9 \\
        1 & \text{if } 1.9 < \text{Creatinine} \leq 4.9 \\
        2 & \text{if } \text{Creatinine} > 4.9
    \end{cases}
\end{align*}
The state $S_t$ is then defined as the tuple $(\text{O2}_t, \text{MAP}_t, \text{GCS}_t, \text{Creat}_t)$. The action space corresponds to all possible combinations of vasopressor and fluid bolus use, each of which can be either $0$ or $1$. There are thus four possible actions: $(0, 0), (0, 1), (1, 0), (1, 1)$.

\paragraph{Reward function}
We define the reward functions as follows:
\begin{align*}
    R_1(S_t) &= 60 - 10 \left( \text{MAP}_t + \text{Creat}_t \right) \\
    R_2(S_t) &= 60 - 10 \left( \text{O2}_t + \text{GCS}_t \right)\\
    R_w(S_t) &= w R_1(S_t) + (1-w) R_2(S_t)
\end{align*}
The reward function $R_1$ incentivizes MAP and Creatinine levels within the desired range, and are proxied as indicators of hypotension and kidney function.
The reward function $R_2$ incentivizes O2 and GCS levels within the desired range, and are proxied as indicators of consciousness and sedation during the ICU stay.
A reward preference of $w=0$ chooses $R_2$ and $w=1$ chooses $R_1$.
We visualize these reward functions for different preferences $w$ in Fig~\ref{fig:fig_mimic_rewards} in the supplement.

\paragraph{Transition model learning}
Since the state and action spaces are discrete, we learn a tabular transition model. This amounts to learning the transition probabilities $P(S_{t+1} | S_t, A_t)$ for all possible state-action pairs.
For the MLE model, we estimate these probabilities by counting the number of times each transition occurs in the training data and normalizing by the total number of transitions from state $S_t$ under action $A_t$ (we also add a pseudocount of $0.01$ to avoid zero probabilities).

\paragraph{Policy Learning}
The tabular transition model allows us to use Value Iteration (by applying Equation \ref{eq:bellman_op}) to learn the optimal policy.
To ensure that the learned policies don't take unsafe actions (i.e. actions not observed in the training data), we force the policies to have zero probability on actions that have less than 3\% probability in the training data.
For the DF and RDF models, we use consistent weighted per-decision importance sampling (CWPDIS) \citep{metelli2020importance} to estimate the return of a policy $\pi(\theta)$ using the training data. The CWPDIS estimator is given by:
\begin{align*}
    \hat{J}^{\text{CWPDIS}}(\pi) &= \sum_{t=1}^{T} \gamma^{t} \frac{\sum_{n \in \mathcal{D}} r_{nt} \rho_{nt}(\pi)}{\sum_{n \in \mathcal{D}} \rho_{nt}(\pi)},
    &\quad \rho_{nt}(\pi) &= \prod_{k=1}^{t} \frac{\pi(a_{nk} | s_{nk})}{\mu(a_{nk} | s_{nk})},
\end{align*}
where $r_{nt}$ is the reward at time $t$ in trajectory $n$, $\rho_{nt}(\pi)$ is the called the importance weight, $\pi(\theta)$ is the policy being evaluated, and $\mu$ is the behavior policy (estimated from the data).

\begin{figure}[t]
    \centering
    \includegraphics[width=.98\linewidth]{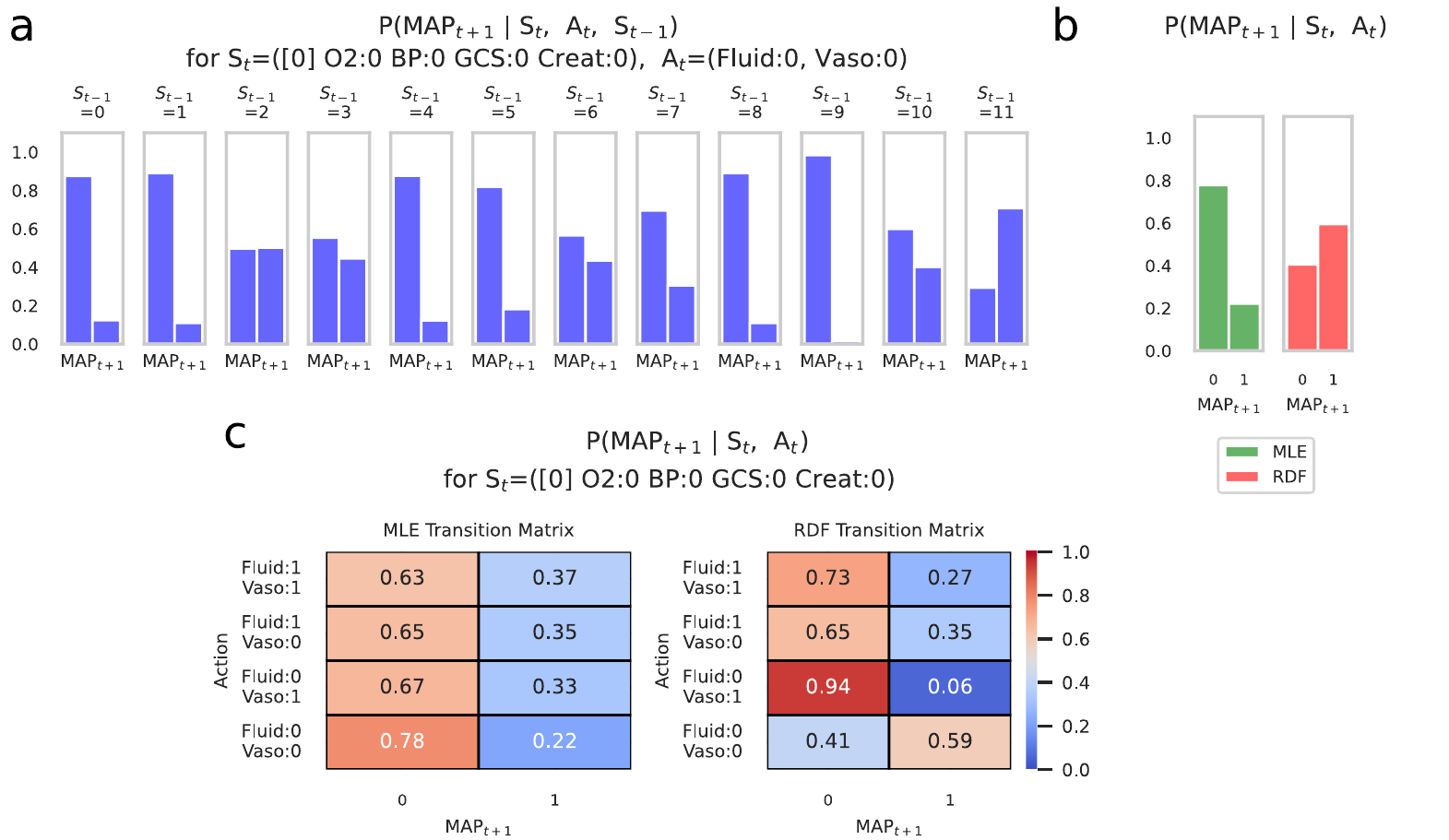}
    \caption{Non-Markovian Transition Dynamics: (a) The dataset suggests that knowing the present state $S_t=0$ and action $A_t=0$ are not sufficient to determine the next state's MAP$_{t+1}$ distribution (i.e. transitions are non-Markovian) since this distribution changes as $S_{t-1}$ is varied.
    (b) MLE and RDF learn different transition dynamics under the constraint that the transition model is Markovian.
    (c) Learned MLE and RDF transition probabilities for the transition MAP$_{t+1} | S_t=0, A_t$ for all actions.
    }
    \label{fig:mimic_nonmarkovian}
\end{figure}
\subsection{Conclusions from ICU Application}
\paragraph{Learning a tabular transition matrix is useful for interpreting what the model learned about the dynamics}
Fig~\ref{fig:rdf_mimic_transitions} shows the transition dynamics learned by the RDF model (DF and MLE models in the appendix).
Choosing a tabular class of transition model helps the practitioner inspect the transition dynamics learned by the model.
Such an inspection is useful as it allows model critique and revision if necessary.
Complex model classes (e.g. NNs) can achieve good predictive performance and decision quality, but their black box nature precludes this ability to inspect the model.
The RDF approach allowed us to use the simpler model classes, while still achieving good decision quality.
\paragraph{RDF model achieves the best transfer to different reward functions}
As seen in the earlier results, the RDF model achieves superior performance both on and away from the learning-phase reward preferences, as demonstrated by Fig~\ref{fig:mimic_results}(a).
There is a significant drop in the performance of the DF transition model, providing more evidence that DF models are prone to severe overfitting to the reward function they were trained with ($w=1$ in this case).
This suggests that DF model does worse for patients who have a GCS less than $12$ or O2 ratio less than $200$.
The MLE model is less sensitive to the reward preference but also has low performance overall.
\paragraph{RDF can adjust for non-Markovian dynamics by learning Markovian transition dynamics useful for decision-making}

Markovian assumption means that the future outcome of a process is independent of the past given the present.
In the context of the MIMIC dataset, this would mean that knowledge of the present present state $S_t$ and action $A_t$ is sufficient to predict the next state $S_{t+1}$, and that the past states and actions $S_{1:t-1}$ and $A_{1:t-1}$ are not necessary for this prediction.
This is clearly a simplifying assumption that does not hold in general. For example, in our dataset, the future MAP severity (MAP$_{t+1}$) of a patient is likely to depend on the past MAP severity (MAP$_{1:t}$) of the patient. Indeed, we observe that the distribution of MAP$_{t+1}$ given $S_t=0$ (healthy state) and $A_t=0$ (no action) depends on whether the patient was in a healthy state ($S_{t-1}=0$) or a very sick state ($S_{t-1}=0$) in the previous time step (Fig~\ref{fig:mimic_nonmarkovian}).
Nevertheless, the Markovian assumption on states is a common simplification in reinforcement learning \citep{futoma_popcorn_2020} and is often used in practice to reduce the complexity of the learning problem. Knowing that we are working with a simplification, we would hope to learn a model which can still learn to make good decisions.
The RDF model learns the most useful dynamics for getting a high-return policy (Fig~\ref{fig:mimic_nonmarkovian}(b and c).
This means it chooses to learn transitions which are opposite of what the average (i.e. MLE) would suggest (Fig~\ref{fig:mimic_nonmarkovian}(b and c)).

\paragraph{RDF can allow high-quality personalized policies for different patient cohorts}
A consequence of the RDF model learning is that for different patient cohorts, we can learn personalized policies that are tailored to the requirements of that cohort.
Consider, for example, the patient cohort with coma (GCS value less than 8). For this cohort, the reward preference of $w=0.25$ is more appropriate than the learning-phase reward preference of $w=1.0$ since it incentivizes GCS values above 12.
Fig~\ref{fig:mimic_results}(b) shows the superior transfer performance of the RDF model for this cohort.
In the left panel of Fig~\ref{fig:mimic_results}(c), we inspect the trajectory of a patient in the coma cohort.
We see that the RDF policy (for $w=0.25$) chooses actions that match the expert policy (behavior) at the point of transition to a healthy GCS state.
We see the same behavior in the right panel of Fig~\ref{fig:mimic_results}(c) for a patient in the non-coma cohort (but high-creatinine cohort).

\section{Discussion and Limitations} 
\label{sec:discussion}

In this paper, we introduced the Robust Decision-focused Model-based RL framework (RDF) as a novel approach for learning transition models in settings with varying reward preferences.
Our RDF framework enabled us to learn transition models in settings with varying reward preferences.

We assumed the existence of a model class that is both interpretable and useful for learning a good decision policy. However, this assumption is often valid in clinical settings: the elements important for making decisions frequently end up being simple. Another potential risk is that we may learn incorrect relationships in the data if there are missing features or rewards. However, avoiding this limitation is possible with the help of collaboration with clinicians, and we expect our method to be especially useful in settings where the clinician can provide the relevant features and the range of rewards relevant to the decision-making task. We prioritize inspectability of the model precisely to enable a clinical expert to identify associations that do not make sense due to missing features.

From a technical perspective, an important aspect was addressing the time and space complexity requirements of the RDF framework.
We demonstrated that our RDF framework can use a number of different policy optimizers.
Settings like the MIMIC Hypotension dataset were small enough such that we could apply Value Iteration to learn the optimal action-value function at each planning step.
In other settings, such as the cancer simulator, we combined our RDF framework with a Linear Quadratic Regulator (LQR) planner to make that planning step more efficient. This is an example of a situation in which using a restricted model class (linear) created computational advantages in addition to being interpretable.

To scale the RDF framework to more reward bases and more complex environments, we can use more sophisticated policy planners, such as a Deep Q-Network (DQN) or a Proximal Policy Optimisation (PPO) planner. For these methods, learning an optimal policy at each planning step is computationally prohibitive. However, we can leverage past work that has shown that these interleaving $Q-$ and $\theta-$ learning steps makes it possible to achieve convergence in decision-focused learning with neural-network-based policy optimizers \citep{nikishin_control-oriented_2022}. To mitigate the computational and space complexity of storing the $Q-$values for each reward preference, we can use a single network.

Our work assumed access to the reward preferences at the learning and deployment stages. While the RDF framework is robust to errors in these preferences by design, future work could explore how to learn these preferences from data.
We also note that decision-focused learning can be used with complex transition model classes, such as neural networks, and that our RDF framework can be readily extended to these settings. We limited our work to simple transition models due to our focus on healthcare applications which often require interpretable models. 
Finally, we note that RDF learning proceeds by learning a separate policy for each reward preference. Future work could explore how to learn a single policy that can adapt to different reward preferences at deployment time. Combining RDF with multi-objective reinforcement learning, which has focused on the development of (mostly) model-free techniques to learn such policies, could be a promising direction for this work.

\subsection{Clinical Applicability}

Our approach could help in cases where multiple objectives are important from a clinical perspective \citep{zhang2022interpretable}, by learning a simple transition model in MDP that focuses on its downstream use of providing policies for multiple rewards (MAP, Mortality probability, and Final Survival in the case of \citet{zhang2022interpretable}).
Our approach can also be beneficial in scenarios where there is legitimate uncertainty about which reward to use. For instance, in the selection of cancer dosing regimens \citep{yauney_reinforcement_2018}, the reward function specification required setting several coefficients.
We adapted this environment in Section \ref{sec:healthcare_experiments} by proposing a reward function that is a weighted combination of (1) the reduction in tumor size and (2) drug concentration.

\acks{
This material is based upon work supported by the National Science Foundation under Grant No. IIS-1750358, Grant No. IIS-2007076 and by NIH award R01MH123804.
Any opinions, findings, and conclusions or recommendations expressed in this material are those of the author(s) and do not necessarily reflect the official views of the National Science Foundation or the National Institutes of Health.}

\bibliography{mybibfile}

\newpage
\appendix

\section{Alternate RDF Objective Formulation}
\label{sec:alt_rdf_formulation}
The RDF objective can also be formulated in the integral terms
\begin{align}
    \label{eq:RDF_appdx}
    \theta_{\RDF} \gets &\arg\min_{\theta} \int_{w \in \mathcal{U}} {L_{T_\ast,R_w}(\theta)} dw  \notag \\
    & \text{s.t. } J_{T_\ast,R_{\bar{w}}}(\pi^\ast(\theta, R_{\bar{w}})) \geq \delta
\end{align}
where $\mathcal{U}$ specifies the region over which we wish to be robust over.

While this objective is equivalent to the objective in Eqn \ref{eq:RDF} under the assumption that $P(w)$ is a uniform measure on the domain of $\mathcal{U}$, it has an intuitive interpretation: we wish to maximize the volume of the return achieved by our model $\theta$. This intuition also motivates why we choose a uniform measure for $P(w)$.




\section{Proof for Theorem 1}
\label{sec:proof}

\begin{theorem} 
    Let $R_{\bar{w}}$ be the learning-phase reward function with preference $\bar{w}$, and $R_w$ be the reward function with an arbitrary preference $w$.
    Let $\Qoptw$ be the optimal action-value function for the true MDP for reward function $R_w$.
    Let $\Boptw, \Bdfw, \Brdfw$ denote the Bellman optimality operators under the true dynamics, DF model and RDF model respectively.
  
    Assume $\Qdfw$ and $\Qrdfw$ are fixed points under $\Bdfw$ and $\Brdfw$ respectively.
    Further assume that the reward function is bounded, $R_w(s,a) \in [0, \rmax] \forall s, a, w$.
    \paragraph{DF Case}
    Consider a DF model trained on $R_{\bar{w}}$. 
    If the Bellman operator induced by the DF model achieves the error
    $$\sup_{s,a} |\Boptwbar \Qdfwbar \sa - \Bdfwbar \Qdfwbar \sa| = \epsdfwbar,$$
    then
    \begin{align}
        \Qoptw \sa - \Qdfw \sa &\leq \frac{\epsdfw}{(1-\gamma)} &\quad \textit{ for $w = \bar{w}$} \\
        \Qoptw \sa - \Qdfw \sa &\leq \gamma\frac{\rmax}{(1-\gamma)^2} &\quad \textit{$\forall w \neq \bar{w}$}
    \end{align}
    \paragraph{RDF Case}
    Consider the RDF model trained with learning-phase preference $\bar{w}$ and deployment-phase reward preference distribution $P(w)$.
    For a $w \in P(w)$, if the Bellman operator $\Brdfw$ induced by the RDF model achieves the error
    $$\sup_{s,a} |\Boptw \Qrdfw \sa - \Brdfw \Qrdfw \sa| = \epsrdfw,$$
    then
    \begin{align}
    \Qoptw \sa - \Qrdfw \sa \leq \frac{\epsrdfw}{(1-\gamma)}
    \end{align}
  \end{theorem}
  For, $w \neq \bar{w}$, the RDF bound is tighter since we explicitly optimize $\epsrdfw$ whereas $\gamma \frac{\rmax}{(1-\gamma)^2}$ is constant.
  
\begin{proof}
First, we show the bound for the RDF and DF($w=\bar{w}$) case. Then we show the DF($w\neq \bar{w}$) bound.

\paragraph*{RDF and DF($w=\bar{w}$) case}
$\forall s, a$ our $Q$ approximation can be written as follows,

\begin{align}
&\left|\Qoptw \sa - \Qrdfw \sa \right| \\
=& \left|\Boptw \Qoptw \sa - \Brdfw \Qrdfw \sa \right| \\
\leq& \left|\Boptw \Qrdfw \sa - \Brdfw \Qrdfw \sa \right| + \left|\Boptw \Qoptw \sa - \Boptw \Qrdfw \sa \right|\\
=& \epsrdfw + \left| r_w \sa + \gamma \mathbb{E}_{T^*(s, a)} \brackets{\max_{a'} \Qoptw (s', a')} - r_w \sa - \gamma \mathbb{E}_{T^*(s, a)} \brackets{\max_{a'} \Qrdfw (s', a')} \right| \\
=& \epsrdfw + \gamma \left|\mathbb{E}_{T^*(s, a)} \brackets{\max_{a'} \Qoptw (s', a') - \max_{a'} \Qrdfw (s', a')} \right| \\
\leq& \epsrdfw + \gamma \norm{T^*(s, a)}_1 \norm{\max_{a'} \Qoptw (\cdot, a') - \max_{a'} \Qrdfw (\cdot, a')}_\infty \\
\leq& \epsrdfw + \gamma \max_{s',a'} \abs{\Qoptw (s', a') - \Qrdfw (s', a')}
\end{align}

Taking a $\max$ over $s, a$ yields
\begin{align}
    \max_{s,a} \left|\Qoptw \sa - \Qrdfw \sa \right| \leq \epsrdfw + \gamma \max_{s,a} \abs{\Qoptw (s, a) - \Qrdfw (s, a)} \\
    \implies \max_{s,a} \left|\Qoptw \sa - \Qrdfw \sa \right| \leq \frac{\epsrdfw}{(1-\gamma)}
\end{align}

For the DF($w = \bar{w}$) case, we get an analogous bound:
\begin{align}
    \max_{s,a} \left|\Qoptwbar \sa - \Qdfwbar \sa \right| \leq \frac{\epsdfwbar}{(1-\gamma)}
\end{align}

\paragraph{DF($w \neq \bar{w}$) case}

First, we show that for any $Q \sa$ and $w \neq \bar{w}$,
\begin{align}
    &\abs{\Boptw Q \sa - \Bdfw Q \sa} \\
    =& \gamma \abs{ \mathbb{E}_{T^*(s, a) - T^{DF}(s, a)} \brackets{\max_{a'} Q (s',a')} } \\
    \leq& \gamma \abs{ \mathbb{E}_{T^*(s, a) - T^{DF}(s, a)} \brackets{\max_{a'} Q (s',a')} - \frac{\rmax}{2(1-\gamma)}} \\
    \leq& \gamma \norm{T^*(s, a) - T^{DF}(s, a)}_1 \norm{\max_{a'} Q (\cdot,a') - \frac{\rmax}{2(1-\gamma)} \mathbf{1} }_\infty \\
    \leq& \gamma \norm{T^*(s, a) - T^{DF}(s, a)}_1 \frac{\rmax}{2(1-\gamma)} \\
    \leq& \gamma \frac{\rmax}{(1-\gamma)}
\end{align}

Now, for a $\sa$ and $w \neq \bar{w}$

\begin{align}
    &\left|\Qoptw \sa - \Qdfw \sa \right| \\
=& \left|\Boptw \Qoptw \sa - \Bdfw \Qdfw \sa \right| \\
\leq& \left|\Boptw \Qdfw \sa - \Bdfw \Qdfw \sa \right| + \left|\Boptw \Qoptw \sa - \Boptw \Qdfw \sa \right|\\
\leq& \left|\Boptw \Qdfw \sa - \Bdfw \Qdfw \sa \right| + \gamma \max_{s',a'} \abs{\Qoptw (s', a') - \Qdfw (s', a')}\\
\leq& \gamma \frac{\rmax}{(1-\gamma)} + \gamma \max_{s',a'} \abs{\Qoptw (s', a') - \Qdfw (s', a')}
\end{align}
which yields
\begin{align}
    \max_{s,a} \left|\Qoptw \sa - \Qdfw \sa \right| \leq \gamma \frac{\rmax}{(1-\gamma)^2}
\end{align}

\end{proof}

\section{Environment Details}

\subsection{Synthetic MDP}
The transition matrix values were sampled from a standard normal distribution before being normalized using a softmax transformation. There are two reward matrices, and their values were uniformly sampled in $[-60, 40]$ before being clipped to $[0,40]$. 

We share the code for generating the synthetic MDP for validating the Theorem 1 in the main paper. The code is written in Python and uses PyTorch.

\begin{lstlisting}[language=Python]
data_seed = 0
torch.manual_seed(data_seed)
num_actions, num_states = 2, 20
dtype_float = torch.float64

# Transition matrix
# Shape: [num_actions, num_states, num_states]
true_transition = torch.randn(num_actions, num_states, num_states, dtype=dtype_float)
true_transition = torch.softmax(true_transition, dim=-1)
true_transition[:,-1] = 0
true_transition[:,-1,-1] = 1

# Reward matrix
# Shape: [num_states, num_actions]
true_reward_1 = (torch.rand(num_states, num_actions, dtype=dtype_float))
true_reward_2 = (torch.rand(num_states, num_actions, dtype=dtype_float))
true_reward_1 = true_reward_1.clamp(0.6,) - 0.6
true_reward_2 = true_reward_2.clamp(0.6,) - 0.6
true_reward_1 *= 100
true_reward_2 *= 100
true_reward_1[-1,:] = 0
true_reward_2[-1,:] = 0
get_true_reward = lambda w: true_reward_1 * w + true_reward_2 * (1 - w)

gamma = 0.9
temperature = 0.01

\end{lstlisting}

\section{Additional Plots}

\subsection{Cancer Treatment}

\subsubsection{Sensitivity analysis for $\lambda$ and $\delta$ parameters}

The optimization problem in Eqn \ref{eq:RDF} is solved by the saddle-point algorithm, where the Lagrange multiplier $\lambda$ is updated depending on whether the constraint is satisfied or not. However, we can also solve the optimization problem by fixing $\lambda$ and running the optimization on a grid of such $\lambda$ values.

\begin{figure}[H]
    \centering
    \begin{minipage}{0.45\linewidth}
        \includegraphics[width=\linewidth]{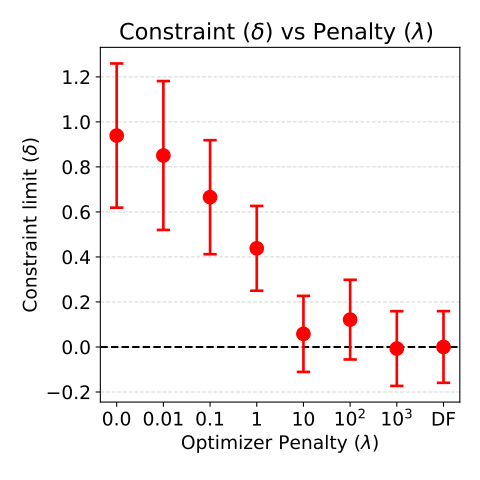}
    \end{minipage}
    \begin{minipage}{0.45\linewidth}
        \includegraphics[width=\linewidth]{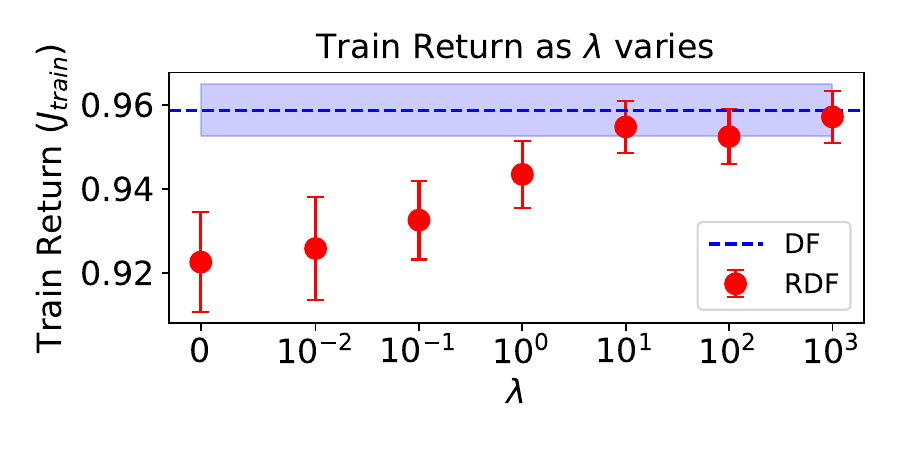}
        \includegraphics[width=\linewidth]{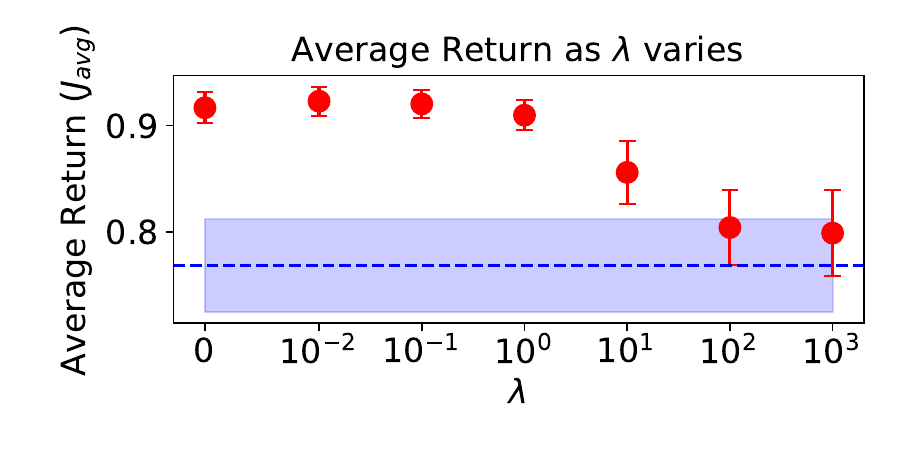}

    \end{minipage}
    \label{sensitivity_lambda}
    \caption{Sensitivity analysis of the $\lambda$ hyperparameter. \textbf{Left:} $\lambda$ vs $\delta$ for the cancer treatment problem. \textbf{Right:} $\lambda$ vs learning-phase (i.e. train) and average return for the cancer treatment problem.
}
\end{figure}

In figure\ref{sensitivity_lambda}, we show the implicit constraint value $\delta$ achieved by the optimization problem for different values of $\lambda$.

\subsubsection{Learned model coefficients for MLE, DF, and RDF models for the Cancer Treatment domain}

\resizebox{\columnwidth}{!}{%
\begin{tabular}{llrrr}
\toprule
 &  & MLE Coefficients & DF Coefficients & RDF Coefficients \\
Outcome & State/Action &  &  &  \\
\midrule
\multirow[c]{7}{*}{Concentration} & Intercept & -0.00 & -0.01 & -0.01 \\
\cline{2-5}
 & Concentration & 0.76 & 0.64 & 0.83 \\
\cline{2-5}
 & Proliferative Tissue & -0.00 & -0.05 & -0.06 \\
\cline{2-5}
 & Non-proliferative Quiescent Tissue & -0.00 & 0.02 & -0.03 \\
\cline{2-5}
 & Damaged Quiescent Cells & -0.00 & -0.10 & -0.01 \\
\cline{2-5}
 & Time Step & 0.00 & -0.10 & 0.13 \\
\cline{2-5}
 & Action & 0.76 & 0.60 & 0.69 \\
\cline{1-5} \cline{2-5}
\multirow[c]{7}{*}{Proliferative Tissue} & Intercept & 0.95 & 1.03 & 1.03 \\
\cline{2-5}
 & Concentration & -0.47 & -0.52 & -0.39 \\
\cline{2-5}
 & Proliferative Tissue & 0.93 & 0.92 & 0.96 \\
\cline{2-5}
 & Non-proliferative Quiescent Tissue & -0.01 & 0.00 & -0.08 \\
\cline{2-5}
 & Damaged Quiescent Cells & -0.02 & -0.11 & 0.04 \\
\cline{2-5}
 & Time Step & 0.01 & -0.07 & 0.07 \\
\cline{2-5}
 & Action & -0.62 & -0.58 & -0.62 \\
\cline{1-5} \cline{2-5}
\multirow[c]{7}{*}{Non-proliferative Quiescent Tissue} & Intercept & 3.02 & 3.00 & 2.51 \\
\cline{2-5}
 & Concentration & -1.76 & -1.78 & -1.75 \\
\cline{2-5}
 & Proliferative Tissue & -0.35 & -0.43 & -0.50 \\
\cline{2-5}
 & Non-proliferative Quiescent Tissue & 0.97 & 0.83 & 0.84 \\
\cline{2-5}
 & Damaged Quiescent Cells & -0.08 & -0.01 & 0.13 \\
\cline{2-5}
 & Time Step & 0.07 & 0.05 & 0.08 \\
\cline{2-5}
 & Action & -2.45 & -2.32 & -2.45 \\
\cline{1-5} \cline{2-5}
\multirow[c]{7}{*}{Damaged Quiescent Cells} & Intercept & -2.66 & -2.66 & -2.17 \\
\cline{2-5}
 & Concentration & 1.36 & 1.24 & 1.33 \\
\cline{2-5}
 & Proliferative Tissue & 0.30 & 0.26 & 0.36 \\
\cline{2-5}
 & Non-proliferative Quiescent Tissue & 0.04 & 0.21 & -0.02 \\
\cline{2-5}
 & Damaged Quiescent Cells & 1.06 & 1.13 & 0.93 \\
\cline{2-5}
 & Time Step & -0.05 & -0.21 & 0.01 \\
\cline{2-5}
 & Action & 1.89 & 1.95 & 2.00 \\
\cline{1-5} \cline{2-5}
\bottomrule
\end{tabular}}

\subsubsection{Mean Tumor Diameter Trajectories for different reward preferences - RDF}

\begin{figure}[H]
    \centering
    \includegraphics[width=.7\linewidth]{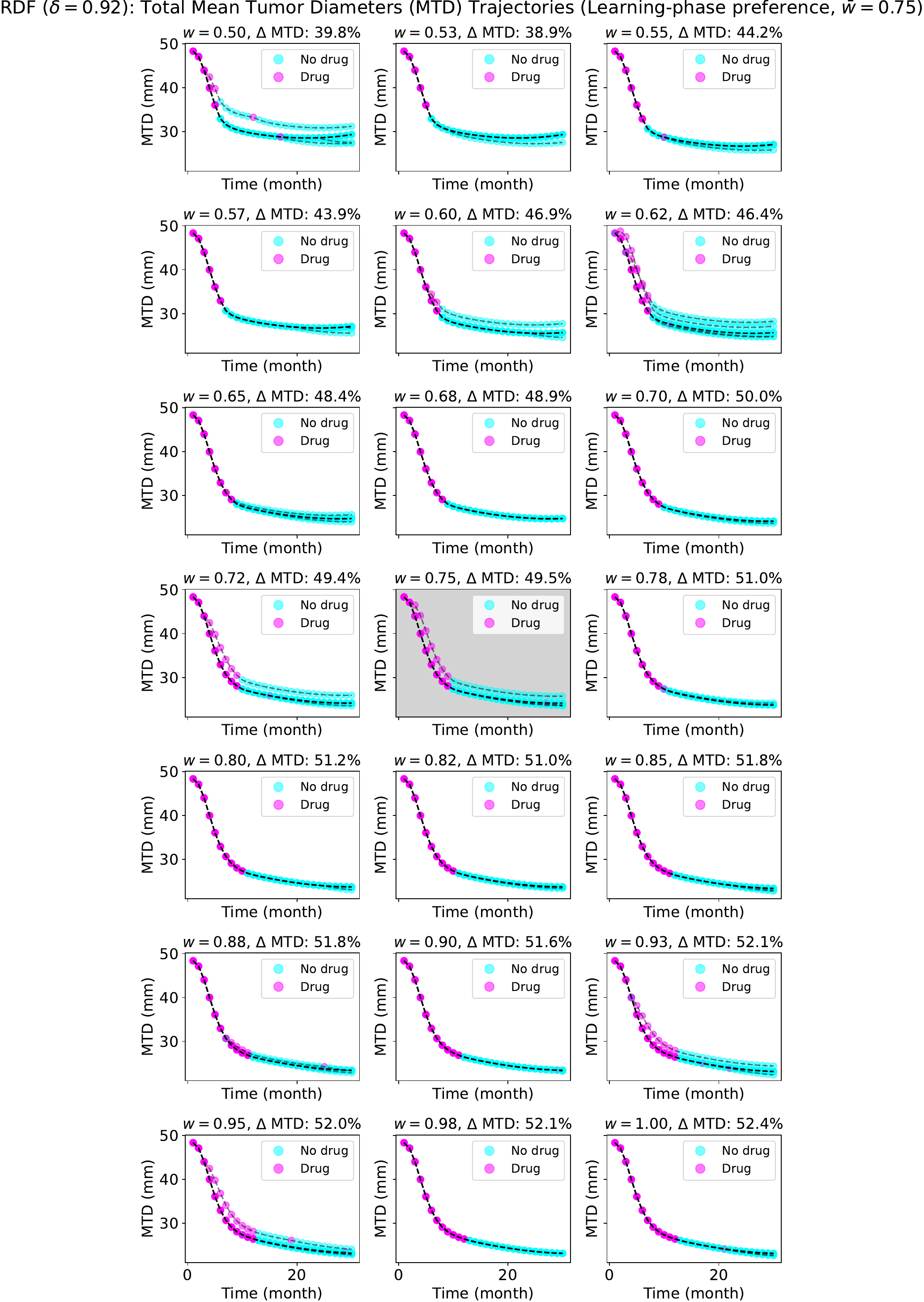}
    \caption{Mean tumor diameter trajectories for different reward preferences with the RDF model. The learning-phase reward preference is $\bar{w} = 0.75$ and the deployment-phase reward preference is $w \in [0.5, 1.0]$.}
\end{figure}

\subsubsection{Mean Tumor Diameter Trajectories for different reward preferences - DF}

\begin{figure}[H]
    \centering
    \includegraphics[width=.7\linewidth]{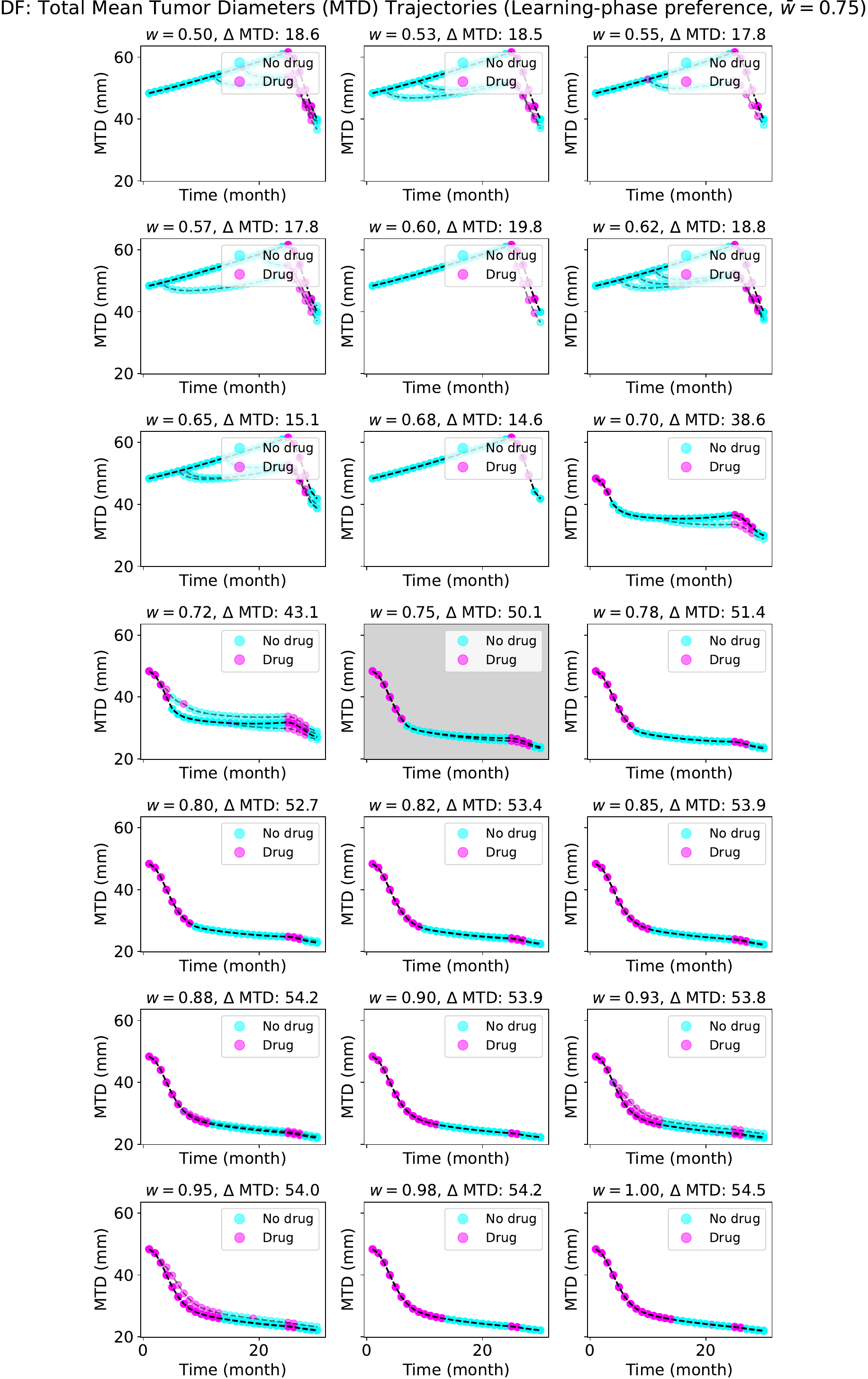}
    \caption{Mean tumor diameter trajectories for different reward preferences with the DF model. The learning-phase reward preference is $\bar{w} = 0.75$ and the deployment-phase reward preference is $w \in [0.5, 1.0]$.}
\end{figure}

\subsubsection{Drug Concentration Trajectories for different reward preferences - RDF}

\begin{figure}[H]
    \centering
    \includegraphics[width=.56\linewidth]{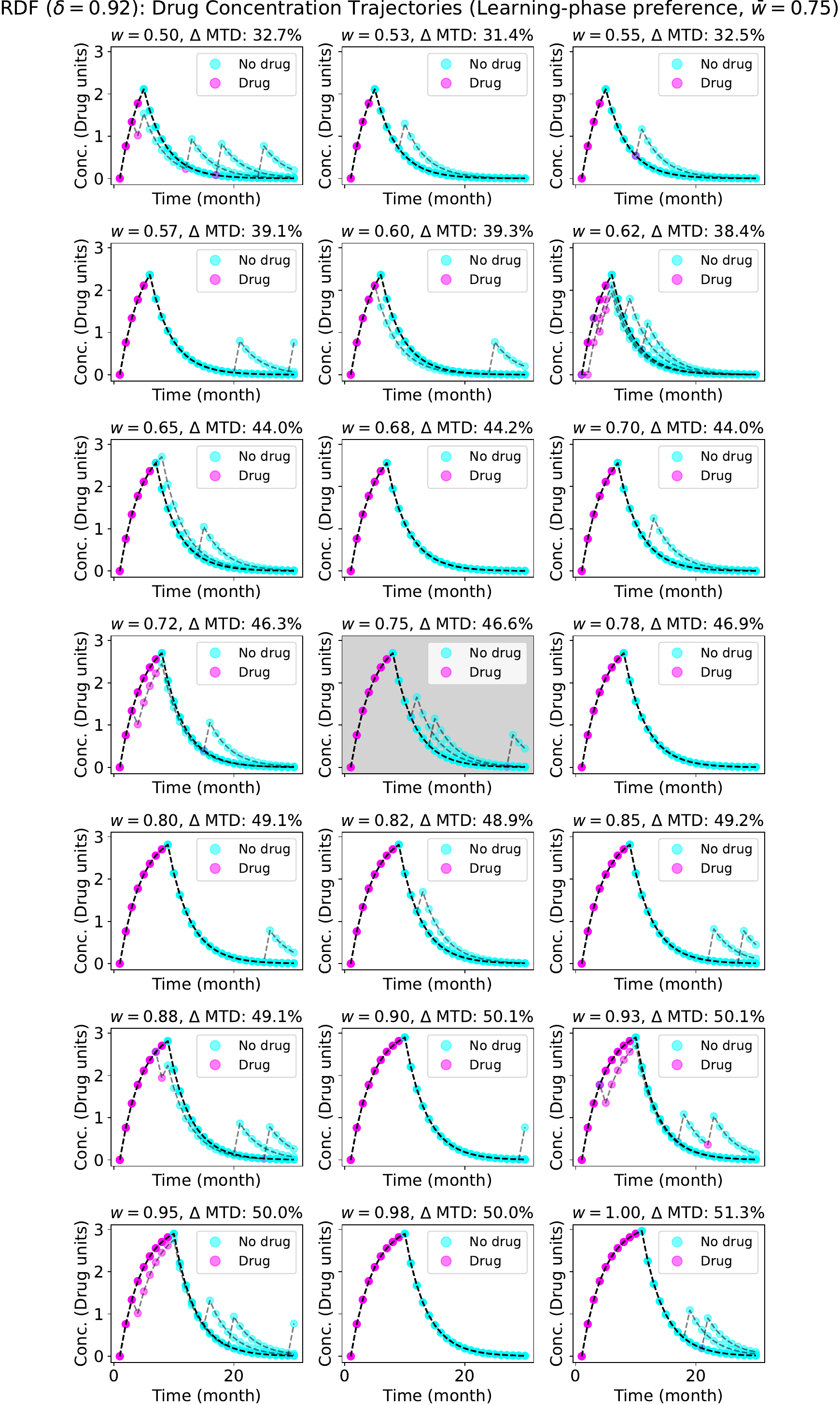}
    \caption{Mean tumor diameter trajectories for different reward preferences with the RDF model. The learning-phase reward preference is $\bar{w} = 0.75$ and the deployment-phase reward preference is $w \in [0.5, 1.0]$.}
\end{figure}

\subsubsection{Drug Concentration Trajectories for different reward preferences - DF}

\begin{figure}[H]
    \centering
    \includegraphics[width=.56\linewidth]{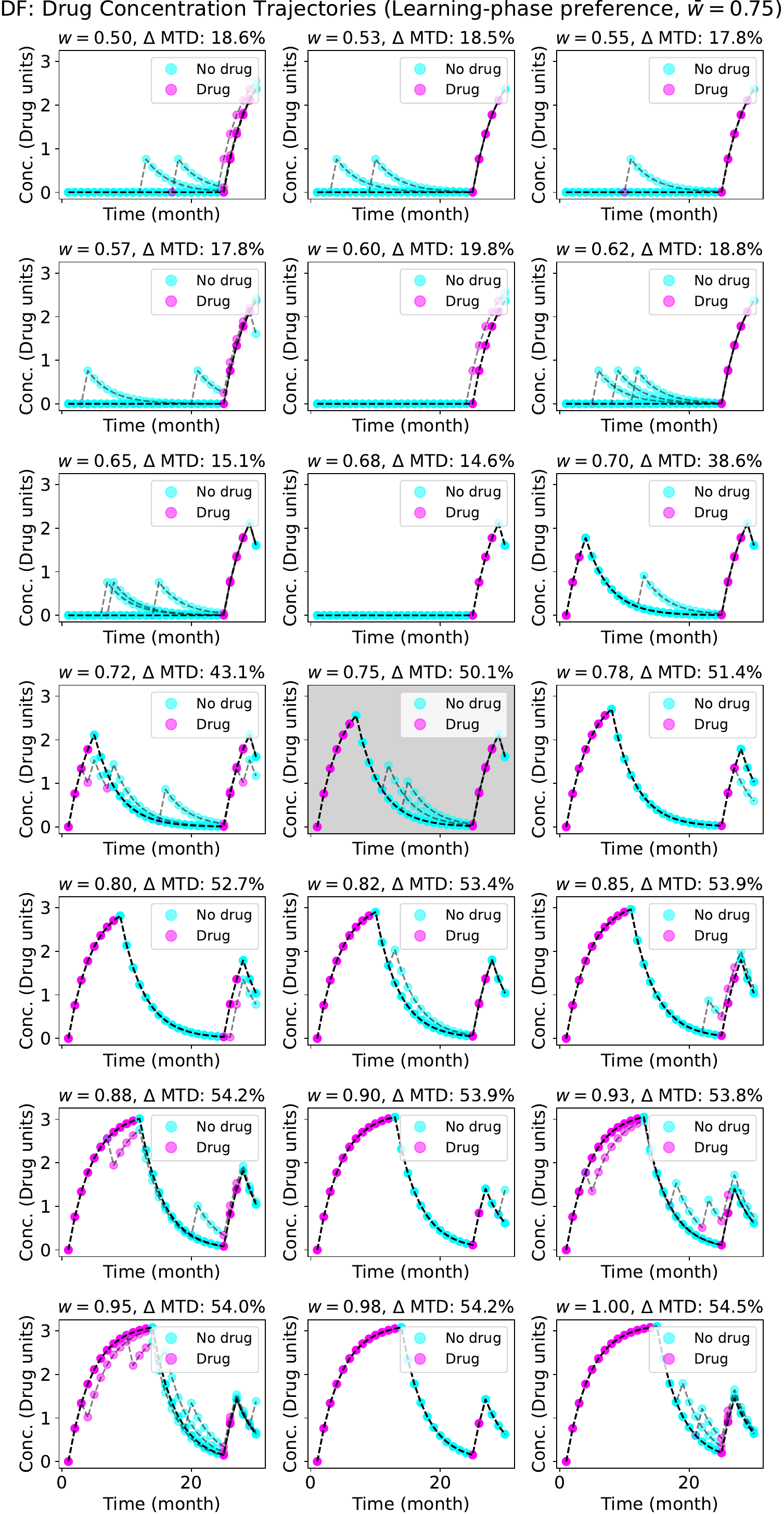}
    \caption{Mean tumor diameter trajectories for different reward preferences with the DF model. The learning-phase reward preference is $\bar{w} = 0.75$ and the deployment-phase reward preference is $w \in [0.5, 1.0]$.}
\end{figure}

\subsection{MIMIC-IV Acute Hypotension dataset}
\label{sec:mimic_appendix}

\begin{figure}[H]
    \begin{minipage}{.9\linewidth}
    \centering
    \includegraphics[width=\linewidth]{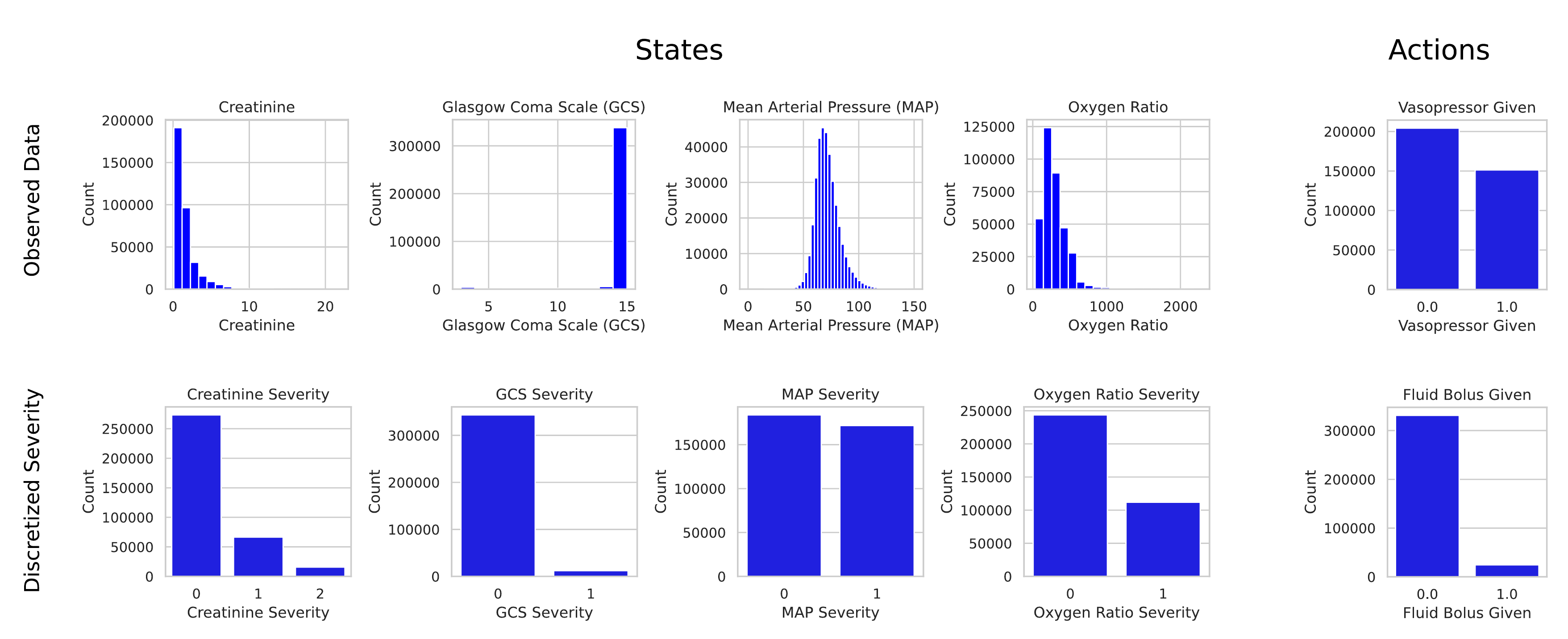}
    \caption{Distributions of states and actions in MIMIC Hypotension dataset}
    \label{fig:fig_mimic_data}
    \end{minipage}
    \hfill
    \begin{minipage}{0.9\linewidth}
    \centering
    \includegraphics[width=\linewidth]{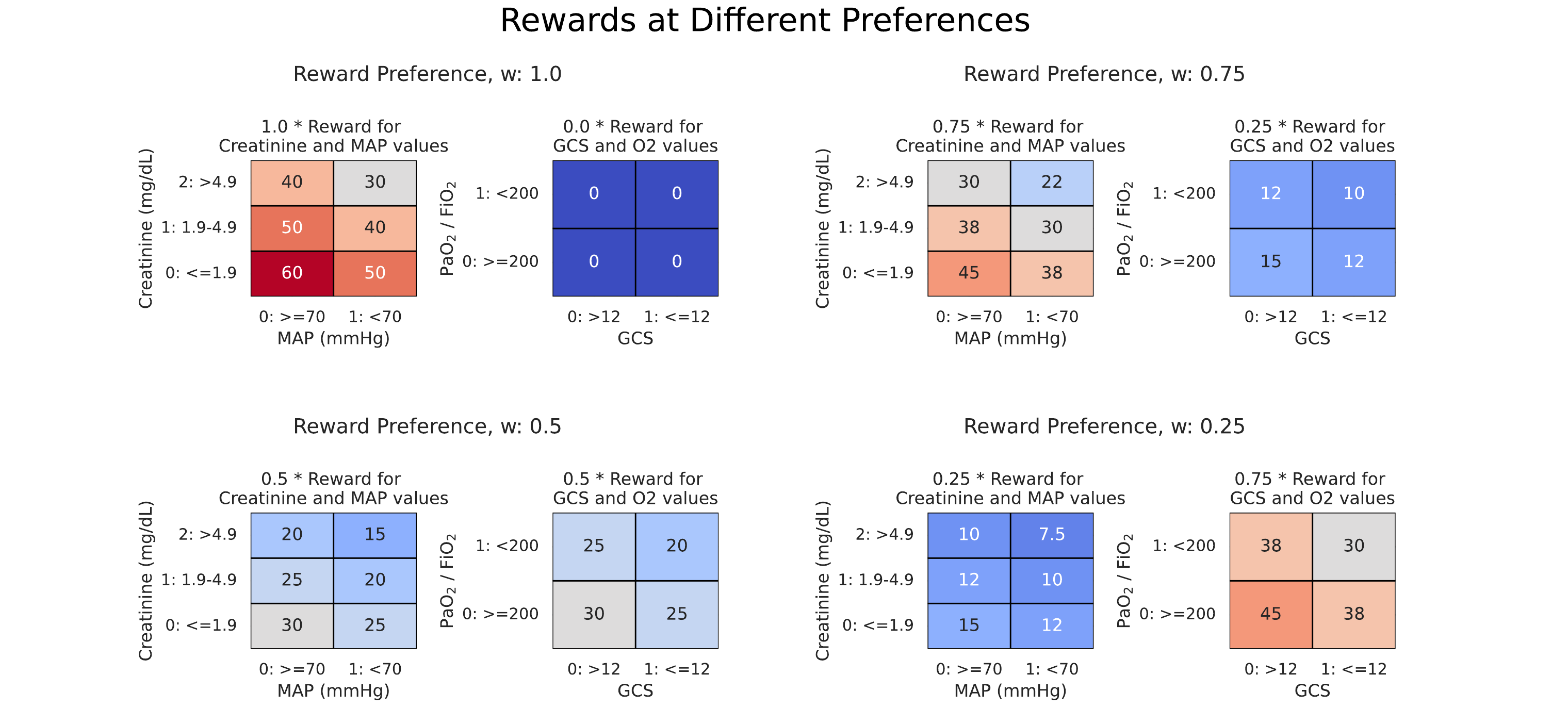}
    \caption{Rewards at different preferences in the MIMIC Hypotension dataset}
    \label{fig:fig_mimic_rewards}
    \end{minipage}
\end{figure}

\subsubsection{Table describing the discretization for the state space}
\begin{table}[H]
\centering
\resizebox{\columnwidth}{!}{%
\begin{tabular}{llcc}
\hline
\textbf{Abbreviation} & \textbf{Clinical Variable} & \textbf{Threshold} & \textbf{Bin Value} \\
\hline
\textbf{O2} & Partial Pressure of Oxygen / Fraction Inspired Oxygen & $\geq 200$ & 0 \\
\textbf{O2} & Partial Pressure of Oxygen / Fraction Inspired Oxygen & $< 200$ & 1 \\
\textbf{BP} & Mean Blood Pressure & $\geq 70$ mmHg & 0 \\
\textbf{BP} & Mean Blood Pressure & $< 70$ mmHg & 1 \\
\textbf{GCS} & Glasgow Coma Scale (GCS) & $\leq 12$ & 0 \\
\textbf{GCS} & Glasgow Coma Scale (GCS) & $> 12$ & 1 \\
\textbf{Crea} & Creatinine & $\leq 1.9$ mg/dL & 0 \\
\textbf{Crea} & Creatinine & $> 1.9$ and $\leq 4.9$ mg/dL & 1 \\
\textbf{Crea} & Creatinine & $> 4.9$ mg/dL & 2 \\
\hline
\end{tabular}}
\caption{Discretization of Clinical Variables into Bins}
\label{table:discretization_bins}
\end{table}

\subsubsection{RDF Transition Dynamics for MIMIC Hypotension data}
\begin{figure}[H]
    \centering
    \includegraphics[width=.98\linewidth]{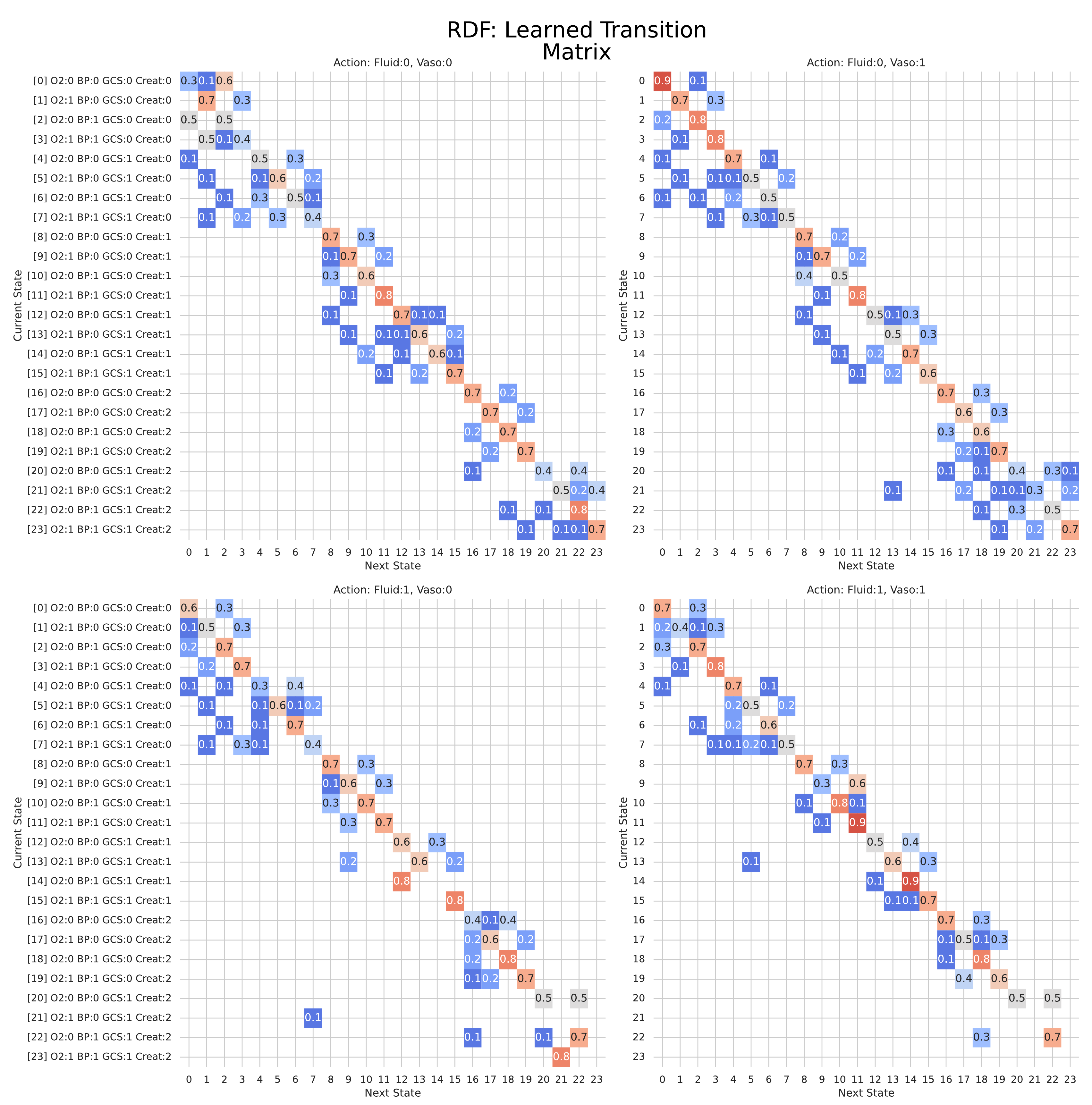}
    \caption{RDF Transition Dynamics for MIMIC Hypotension data.}
    \label{fig:rdf_mimic_transitions}
\end{figure}

\subsubsection{DF Transition Dynamics for MIMIC Hypotension data}
\begin{figure}[H]
    \centering
    \includegraphics[width=.98\linewidth]{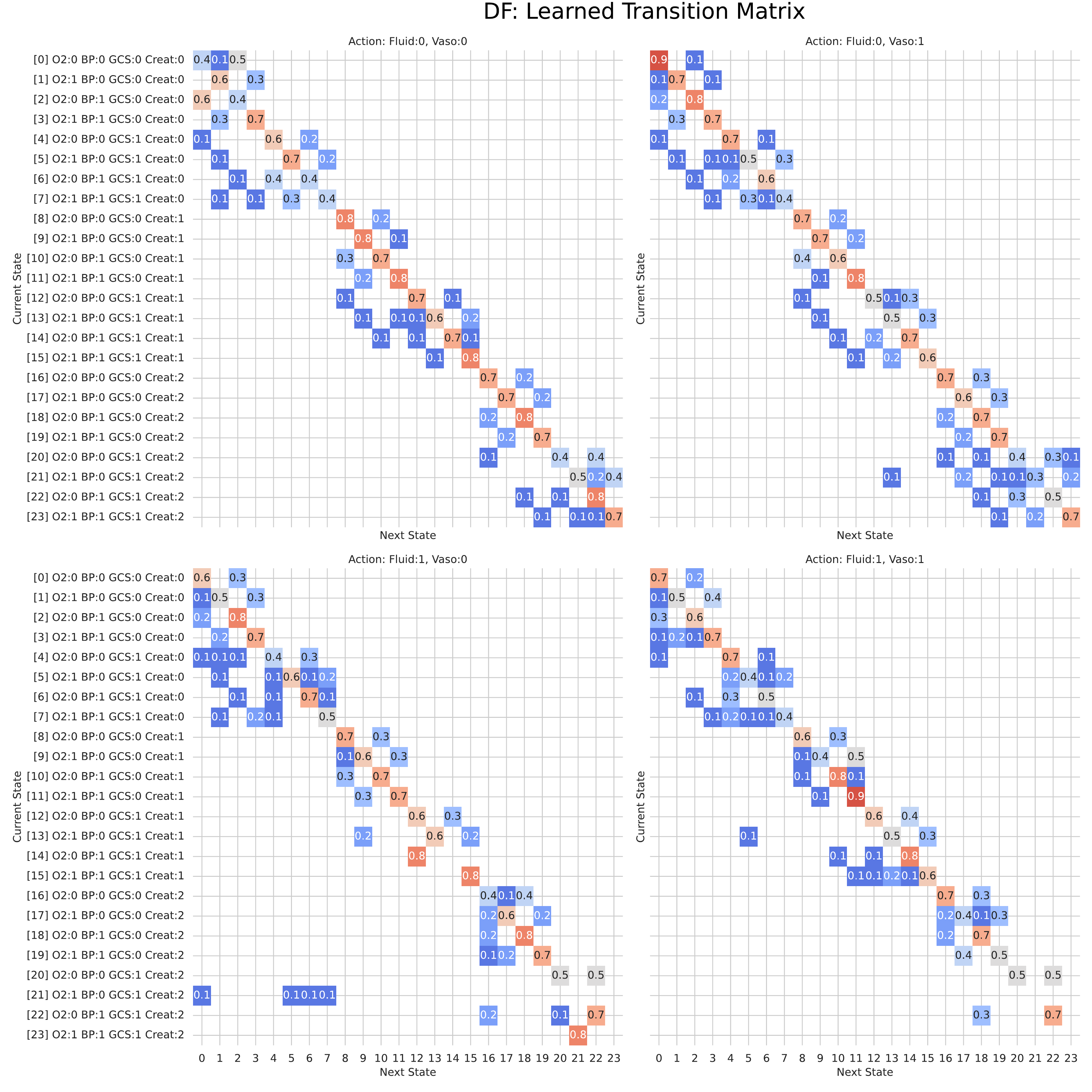}
    \caption{DF Transition Dynamics for MIMIC Hypotension data.}
    \label{fig:df_mimic_transitions}
\end{figure}

\subsubsection{MLE Transition Dynamics for MIMIC Hypotension data}
\begin{figure}[H]
    \centering
    \includegraphics[width=.98\linewidth]{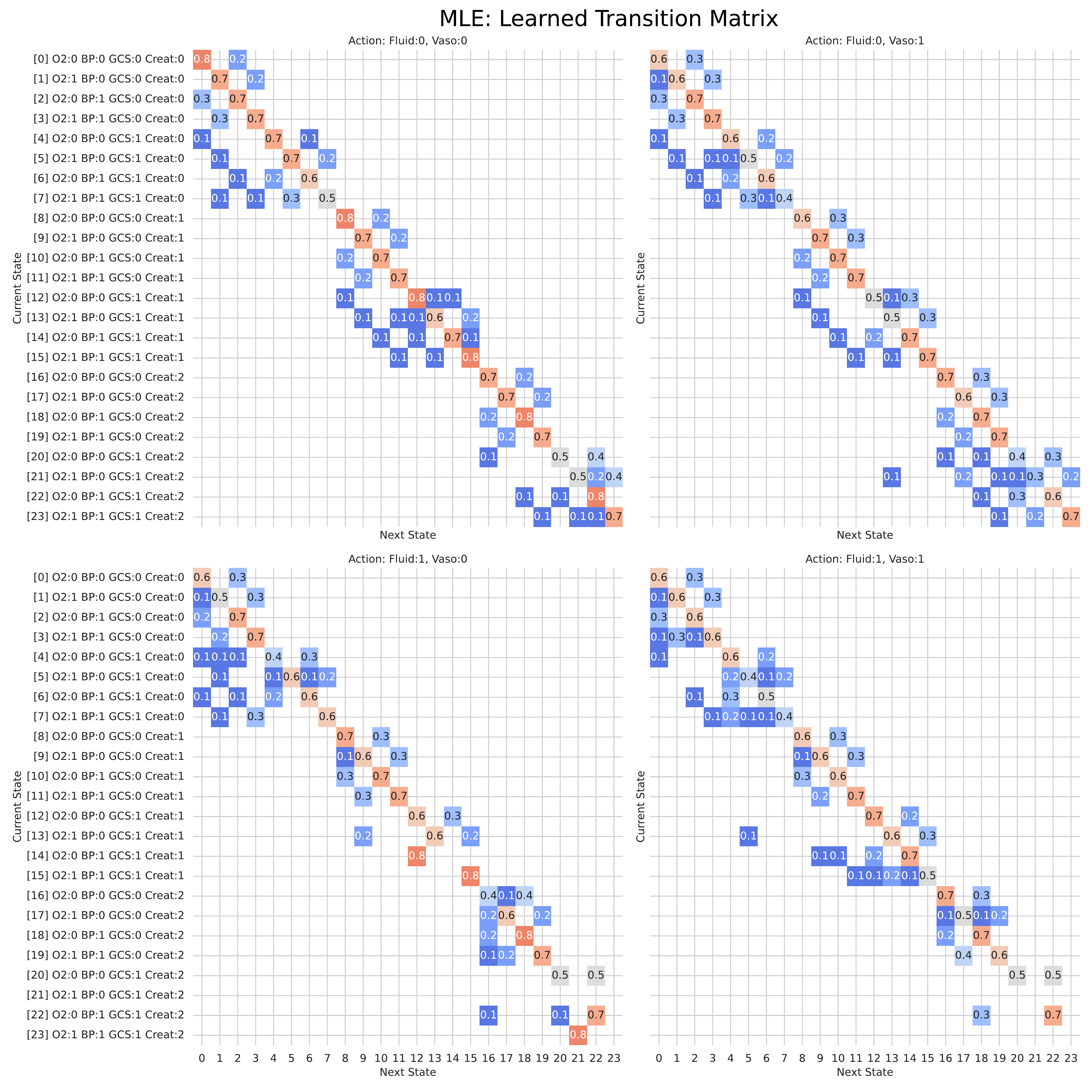}
    \caption{MLE Transition Dynamics for MIMIC Hypotension data.}
    \label{fig:mle_mimic_transitions}
\end{figure}

\end{document}